\documentclass{article}
\PassOptionsToPackage{numbers, compress}{natbib}

\usepackage[preprint]{neurips_2026}

\usepackage[utf8]{inputenc} % allow utf-8 input
\usepackage[T1]{fontenc}    % use 8-bit T1 fonts
\usepackage{hyperref}       % hyperlinks
\usepackage{url}            % simple URL typesetting
\usepackage{booktabs}       % professional-quality tables
\usepackage{amsfonts}       % blackboard math symbols
\usepackage{nicefrac}       % compact symbols for 1/2, etc.
\usepackage{microtype}      % microtypography
\usepackage{xcolor}         % colors

\usepackage{threeparttable}
\usepackage{graphicx}
\usepackage{booktabs}
\usepackage{subcaption}
\usepackage{multirow}
\usepackage{makecell}
\usepackage{array}
\usepackage{tabularx}
\usepackage{amsmath}
\newcolumntype{L}[1]{>{\raggedright\arraybackslash}p{#1}}
\newcolumntype{Y}{>{\raggedright\arraybackslash}X}
\usepackage{pifont}
\newcommand{\cmark}{\ding{51}}
\newcommand{\xmark}{\ding{55}}
\newcommand{\pmark}{\ensuremath{\triangle}}
\usepackage[most]{tcolorbox}
\usepackage{wrapfig}
\usepackage{float}
\usepackage{placeins}

\title{PanoWorld: Towards Spatial Supersensing \\ in 360$^\circ$ Panorama World}

\author{
  \textbf{Changpeng Wang}$^{1}$, \textbf{Xin Lin}$^{2}$, \textbf{Junhan Liu}$^{1}$, \textbf{Yuheng Liu}$^{3}$, \\
  \textbf{Zhen Wang}$^{1}$, \textbf{Donglian Qi}$^{1}$, \textbf{Yunfeng Yan}$^{1}$, \textbf{Xi Chen}$^{4}$ \\[1ex]
  $^{1}$Zhejiang University \quad $^{2}$University of California, San Diego \\
  $^{3}$University of California, Irvine \quad $^{4}$The University of Hong Kong \\
}

\begin{document}

\maketitle

\vspace{-1.5em} 
\begin{center}
    \centering
    \captionsetup{type=figure}
    \includegraphics[width=1\textwidth]{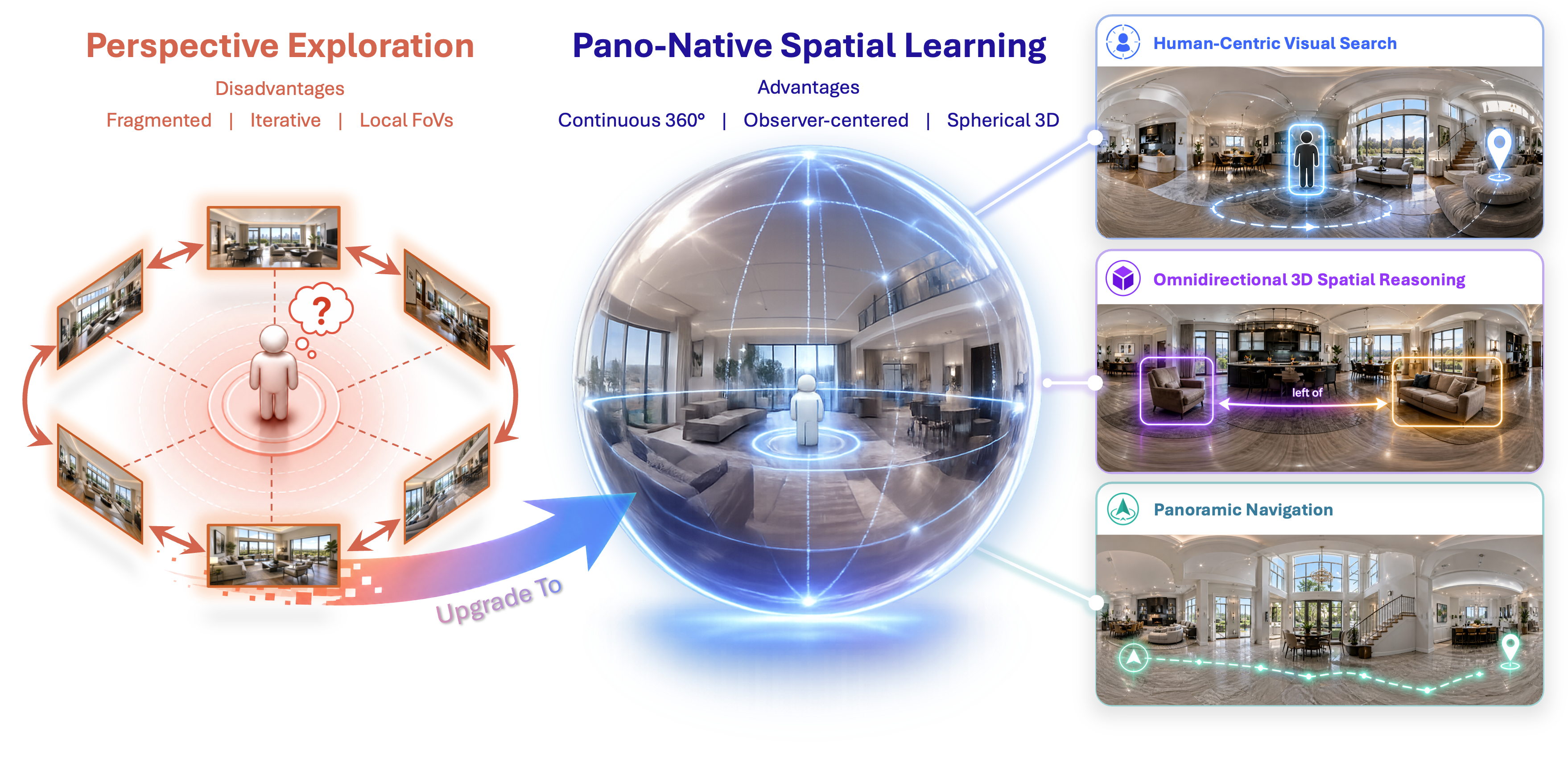} 
    \caption{
    % Can MLLMs reason natively over continuous, observer-centered omnidirectional 3D space? 
    % We shift panoramic reasoning from fragmented perspective-view exploration to pano-native \textbf{supersensing}, enabling 360$^\circ$ spatial reasoning across spherical relations and 3D spatial understanding.
    % Previous MLLM spatial reasoning is largely built on perspective images, requiring models to connect fragmented local views to infer the surrounding space. 
    Existing MLLMs reason over fragmented local views, making it difficult to associate spatial cues in 360$^\circ$. 
    We introduce pano-native supersensing, which teaches VLMs to perceive and reason directly over 360$^\circ$ panoramas, providing a unified full-surround representation for downstream tasks such as human-centric visual search, omnidirectional 3D spatial reasoning, and panoramic navigation.
    }
    \label{fig:teaser}
\end{center}
\vspace{1em}

\begin{abstract}

Multimodal large language models (MLLMs) still struggle with spatial understanding under the dominant perspective-image paradigm, which inherits the narrow field of view of human-like perception.
% where narrow local fields of view make cross-view reasoning difficult.
For navigation, robotic search, and 3D scene understanding, 360$^\circ$ panoramic sensing offers a form of supersensing by capturing the entire surrounding environment at once.
However, existing MLLM pipelines typically decompose panoramas into multiple perspective views, leaving the spherical structure of equirectangular projection (ERP) largely implicit. 
In this paper, we study pano-native understanding, which requires an MLLM to reason over an ERP panorama as a continuous, observer-centered space. 
To this end, we first define the key abilities for pano-native understanding, including semantic anchoring, spherical localization, reference-frame transformation, and depth-aware 3D spatial reasoning.
We then build a large-scale metadata construction pipeline that converts mixed-source ERP panoramas into geometry-aware, language-grounded, and depth-aware supervision, and instantiate these signals as capability-aligned instruction tuning data. 
On the model side, we introduce PanoWorld with Spherical Spatial Cross-Attention, which injects spherical geometry into the visual stream. 
We further construct PanoSpace-Bench, a diagnostic benchmark for evaluating ERP-native spatial reasoning. 
Experiments show that PanoWorld substantially outperforms both proprietary and open-source baselines on PanoSpace-Bench, H$^\ast$Bench, and R2R-CE Val-Unseen benchmarks.
These results demonstrate that robust panoramic reasoning requires dedicated pano-native supervision and geometry-aware model adaptation. 
All source code and proposed data will be publicly released at \url{https://wcpcp.github.io/PanoWorld}.
% All source code and proposed data will be publicly released.
% raising the Qwen3.5 baseline from 30.8 to 56.5, and transfers effectively to H$^\ast$Bench, where it achieves 56.1 zero-shot and 70.0 after fine-tuning, performs better than the previous model of 38.40. 

\end{abstract}

\vspace{-1em}
\section{Introduction}
\vspace{-0.5em}

% MLLM去做空间理解可以帮助人们去***，能够很好做一些下游，比如embodied等等。一些先前的工作***，其中think in 360考虑了全景具有**优势，包含360场景等等，所以他们在全向3D全景场景来做空间的理解。

% 但是他们这种方案还是输入各种投影的透视图来做理解，尽管类似的工作***将全景弄到透视来做可以利用好现有透视的pretrained大模型高泛化能力。但是**也分析讨论了这类方案的缺陷，比如**gap太大，不够efficient等等，所以他们设计了专门全景的模型。而对于mllm理解，还正是需要突破的关口。

% 为了设计一款专门针对全景空间能力理解的评估系统。我们首先系统分析总结了在全景的全向3D场景中，模型应该具有哪些能力。针对这些能力，我们构建了培养模型对应能力的数据以及对应的评价benchmark。进而，我们也对全景设计了专门的MLLM模型。

% % 我们的贡献可以分为：（1）系统分析了问题，总结了不同的能力并且根据这些能力去设计问题; (2) 我们收集并构造了大量的数据集，来作为模型学习那写能力的基础，并提出了大benchmark测试（3）我们针对全景特性的设计了模型；（4）大量实验证明，我们方案在我们bench上最强，并且zero-shot到think in 360上能力也强，训一个epoch直接是他的二倍

% Recent multimodal large language models (MLLMs) have achieved strong performance in basic visual understanding, yet robust spatial reasoning remains challenging under the perspective-image paradigm, where narrow local fields of view limit full-scene awareness and make 3D relations, spatial memory, and layout understanding difficult~\citep{tong2024cambrian,yang2025thinking,stogiannidis2025mindthegap}.
%
Recent multimodal large language models (MLLMs) have made substantial progress in perspective-image visual understanding, yet robust spatial reasoning remains challenging~\citep{wang2025vl,tong2024cambrian,yang2025thinking,stogiannidis2025mindthegap}.
A key limitation of this paradigm is that it inherits the limited instantaneous field of view of human-like perception, whereas tasks such as human-centric visual search, navigation, and immersive scene understanding benefit from full-surround environmental awareness.
360$^\circ$ panoramic sensing therefore offers a form of supersensing, expanding spatial perception from local views to the entire observer-centered environment.
Despite this potential, current approaches often address full-surround reasoning through sequential perspective exploration, as highlighted by recent efforts~\cite{yu2025thinking360}, 
where a continuous panorama is decomposed into local perspective views to simulate human-like exploration of the surrounding 3D environment.
% which fragments a continuous panorama into local views and forces the model to associate objects, directions, and spatial relations across viewpoints.
%
This naturally raises the question of whether 360$^\circ$ spatial reasoning can be modeled more directly and efficiently from panoramic representations themselves, which encode globally consistent observer-centered scenes, including wrap-around continuity and viewpoint-dependent spatial relations.

However, directly transferring existing perspective models to panoramic understanding is challenging, since panoramic and perspective images exhibit fundamental representation gaps, including geometric distortion, non-uniform spatial sampling, and boundary discontinuities~\cite{lin2025one}.
Although some existing approaches in other panoramic tasks utilize perspective-transfer pipelines through projection and stitching~\cite{cao2025panda, wang2024depth, feng2025dit360, liu2026omniroam, zhong2025omnisam}, 
recent progress suggests that panorama-specific models trained on large-scale panoramic data can achieve stronger performance~\cite{lin2025depth, feng2025dit360}, which highlights the importance of large-scale panoramic supervision for learning pano-native representations. 
These observations suggest that panoramic MLLMs should likewise move beyond perspective transfer toward panorama-specific modeling.
Moreover, to enable a unified MLLM capable of handling diverse spatial reasoning tasks, it is essential to systematically characterize the capabilities. However, existing efforts on panoramic MLLMs are typically organized around individual tasks or benchmarks~\citep{zhou2025dense360, zhang2025r1, yang2025odibench, dongfang2025multimodallargelanguagemodels}, and thus remain fragmented and incomplete, lacking both a systematic understanding of the required capabilities and a unified benchmark to evaluate them.
%
% Consequently, current MLLMs still struggle with panoramic reasoning tasks that require global layout understanding, directional relationships, and viewpoint transformations beyond local observations. 
% \yh{give more references.}

To address these limitations, we propose a unified pano-native spatial learning framework for MLLMs that learns an observer-centered representation of the 360$^\circ$ surrounding environment. 
We begin by introducing a capability-based formulation of panoramic spatial understanding, decomposing it into key components such as spherical localization, relative direction reasoning, viewpoint transformation, 3D spatial relations, and global scene topology. 
Building on this formulation, we develop a metadata-driven pipeline for scalable panoramic data construction and build a comprehensive benchmark that evaluates these capabilities beyond conventional VQA-style metrics. 
As summarized in Table~\ref{tab:data_comparison}, the resulting resource covers 570K ERP panoramas and provides a combination of depth-aware signals, entity-level metadata, scalable annotation, and verified graph supervision not jointly available in prior panoramic resources.
We further introduce a pano-aware MLLM model \textbf{PanoWorld} with Spherical Spatial Cross-Attention, enabling the model to align visual features with the underlying geometry of panoramic inputs. Extensive experiments show that our approach achieves strong performance on the proposed benchmark and generalizes effectively to existing 360$^\circ$ reasoning benchmark H$^\ast$Bench~\cite{yu2025thinking360} and VLN benchmark R2R-CE Val-Unseen~\cite{krantz2020navgraphvisionandlanguagenavigationcontinuous}, substantially outperforming existing methods. 

In summary, our main contributions are as follows:
\vspace{-0.5em}
\begin{itemize}
    \item We systematically formulate panoramic spatial reasoning in MLLMs as a capability-structured problem, and derive a taxonomy including spherical localization, relative direction reasoning, viewpoint transformation, 3D spatial relations, and global scene topology.
    \item Based on this formulation, we develop a scalable metadata-driven pipeline for large-scale panoramic data construction, together with a comprehensive benchmark that systematically evaluates the defined spatial reasoning capabilities.
    \item We propose a pano-aware MLLM model \textbf{PanoWorld} with spherical spatial cross-attention for geometry-consistent panoramic understanding.
    \item Extensive experiments demonstrate that the proposed model has achieved competitive performance on the proposed benchmark and effective transfer to existing 360$^\circ$ reasoning and VLN benchmarks, with substantial gains over prior methods.
\end{itemize}

\vspace{-0.5em}
\section{Related Work}
\vspace{-0.5em}

\begin{table*}[t]
\centering
\scriptsize
\setlength{\tabcolsep}{4.0pt}
\renewcommand{\arraystretch}{1.12}
\caption{
Comparison with representative panoramic resources.
We report the number of underlying panoramic images rather than only QA counts.
\cmark: available; \xmark: unavailable; \pmark: partially available.
}
\label{tab:data_comparison}
\begin{tabular*}{\textwidth}{@{\extracolsep{\fill}} L{0.23\textwidth} c c c c c c c c @{}}
\toprule
\textbf{Resource}
& \textbf{Number}
& \textbf{View}
& \textbf{Scene}
& \makecell{\textbf{Depth}\\\textbf{/ 3D}}
& \makecell{\textbf{Entity}\\\textbf{metadata}}
& \makecell{\textbf{Scalable}\\\textbf{annotation}}
& \makecell{\textbf{QA}\\\textbf{/ Captions}}
& \makecell{\textbf{Verified}\\\textbf{graph}} \\
\midrule

% DA$^2$~\citep{li2025depth}
% & 607K & Pano+Persp. & Mixed
% & \cmark & \xmark & \pmark & \xmark & \xmark \\

% DAP~\citep{lin2025depth}
% & 2M & Pano+Persp. & Mixed
% & \cmark & \xmark & \pmark & \xmark & \xmark \\

PanoCity~\citep{guo2026panovggtfeedforward3dreconstruction}
& 120K & Pano & Outdoor
& \cmark & \xmark & \pmark & \xmark & \xmark \\

Pano-AVQA~\citep{yun2021panoavqagroundedaudiovisualquestion}
& 5.4K & Video & Mixed
& \pmark & \pmark & \xmark & \cmark & \xmark \\

Dense360~\citep{zhou2025dense360}
& 160K & Pano & Mixed
& \xmark & \cmark & \cmark & \cmark & \pmark \\

OmniVQA~\citep{zhang2025r1}
& 1.2K & Pano & Indoor
& \xmark & \pmark & \pmark & \cmark & \xmark \\

ODI-Bench~\citep{yang2025odibench}
& 2K & Pano & Mixed
& \xmark & \pmark & \xmark & \cmark & \xmark \\

CFpano~\citep{zhang2025cfpano}
& 2.7K & Multi-Pano & Mixed
& \xmark & \pmark & \xmark & \cmark & \xmark \\

PanoVQA~\citep{fan2026panovqa}
& 44.6K & Pano & Outdoor
& \xmark & \pmark & \pmark & \cmark & \xmark \\

OSR-Bench~\citep{dongfang2025multimodallargelanguagemodels}
& 4.1K & Pano & Indoor
& \pmark & \pmark & \xmark & \cmark & \xmark \\

PanoEnv~\citep{liu2026panoenv}
& 595 & Multi-Persp. & Mixed
& \cmark & \pmark & \xmark & \cmark & \xmark \\

\midrule
\textbf{PanoWorld}
& \textbf{570K} & \textbf{Pano} & \textbf{Mixed}
& \cmark & \cmark & \cmark & \cmark & \cmark \\
\bottomrule
\end{tabular*}
\end{table*}

\textbf{Pano-Native Panoramic Designing.} 
% 我们认为全景相关的设计一般可以分为数据和模型两个层面。
% 数据上有**在**任务上提出大量数据和benchmark
% 专门设计上***用了loss，**用了球面编码，**用了adpative map
To bridge gaps~\citep{lin2025one} between panoramic and perspective understanding, recent studies have emphasized the need for panorama-specific design. 
Existing efforts mainly address this problem from two perspectives: data and models. 
On the data side, recent works construct task-specific datasets and benchmarks for perception~\cite{lin2025depth, ge2025airsim360, zheng2020structured3d}, and MLLM understanding~\citep{zhou2025dense360, zhang2025r1, yang2025odibench, liu2026panoenv}
On the model side, prior studies introduce pano-aware designs, including distortion map \cite{lin2025depth, deng2021lau, yu2023osrt, coors2018spherenet} and spherical positional modeling~\citep{guo2026panovggtfeedforward3dreconstruction, li2025depth, shen2022panoformer, ling2023panoswin, feng2025dit360, fan2026panovqa}. However, most existing efforts remain centered on specific tasks or designs, rather than a unified capability-level formulation of panoramic MLLM understanding.

\textbf{Spatial Reasoning in Multimodal Large Language Models.}
Recent surveys identify spatial reasoning as a systematic bottleneck for large multimodal models, spanning spatial relations, 3D scene understanding, embodied interaction, and geometry-aware representation learning ~\citep{liu2025spatialsurvey,zheng2025multimodalspatial,zha2025enable3d}. 
Following this perspective, spatial reasoning has become an important axis of multimodal evaluation, covering 2D/3D relations, depth order, relative distance, egocentric memory, and embodied question answering ~\citep{tong2024cambrian,yang2025thinking,majumdar2024openeqa,stogiannidis2025mindthegap,yang2025cambrians,huang2024leo,hong2023threedllm,ma2023sqa3d,cheng2024spatialrgpt}. 
Recent methods therefore introduce spatial supervision or geometry-aware representations, such as depth-aware region features, 3D position embeddings, position-aware video representations, and structured 3D scene tokens ~\citep{chen2024spatialvlm,cheng2024spatialrgpt,zhu2025llava3d,zheng2025video3dllm,wang2025dynam3d}. 
These works demonstrate the value of spatial representation learning, but they primarily rely on perspective images, egocentric videos, multi-view observations, or explicit 3D geometry.

Recently, Thinking in 360 ~\citep{yu2025thinking360} studies human-centric visual search in immersive 360$^\circ$ environments. 
This view-based formulation is natural for embodied visual search, yet it treats the panorama primarily as a source of discrete views and leaves panorama geometry implicit. 
In contrast, we ask whether the ERP panorama itself can serve as the native spatial representation of the surrounding space. 
This motivates our pano-native framework, which injects spherical geometry into ERP visual tokens and trains MLLMs to reason directly over continuous, observer-centered panoramic space.

% Thinking in 360 ~\citep{yu2025thinking360} studies humanoid object and path search in immersive 360-degree environments. 
% It represents the environment with panoramic imagery, but operationalizes search by actively rotating the viewing direction and reasoning over local perspective observations. 
% This view-based formulation is natural for embodied visual search, yet it treats the panorama primarily as a source of discrete views and leaves ERP geometry implicit. 
% In contrast, we ask whether the ERP image itself can serve as the model's native spatial representation. 
% This motivates our ERP-native framework, which injects spherical geometry into visual tokens and trains MLLMs to reason over the continuous, observer-centered space encoded by the panorama itself.

% \input{section/methods}

\vspace{-0.5em}
\section{Method}
\vspace{-0.5em}

Our goal is to enable MLLMs to understand panoramas as continuous, observer-centered 360$^\circ$ spaces. 
We first introduce the geometric preliminaries and task settings in Sec.~\ref{3.1}.
We then present a capability taxonomy for pano-native understanding in Sec.~\ref{3.2}.
Based on the formulation, we describe our large-scale metadata construction pipeline in Sec.~\ref{3.3}. Finally, we introduce our pano-aware MLLM model in Sec.~\ref{3.4}.

\begin{figure*}[t]
    \centering
    \includegraphics[width=0.98\textwidth]{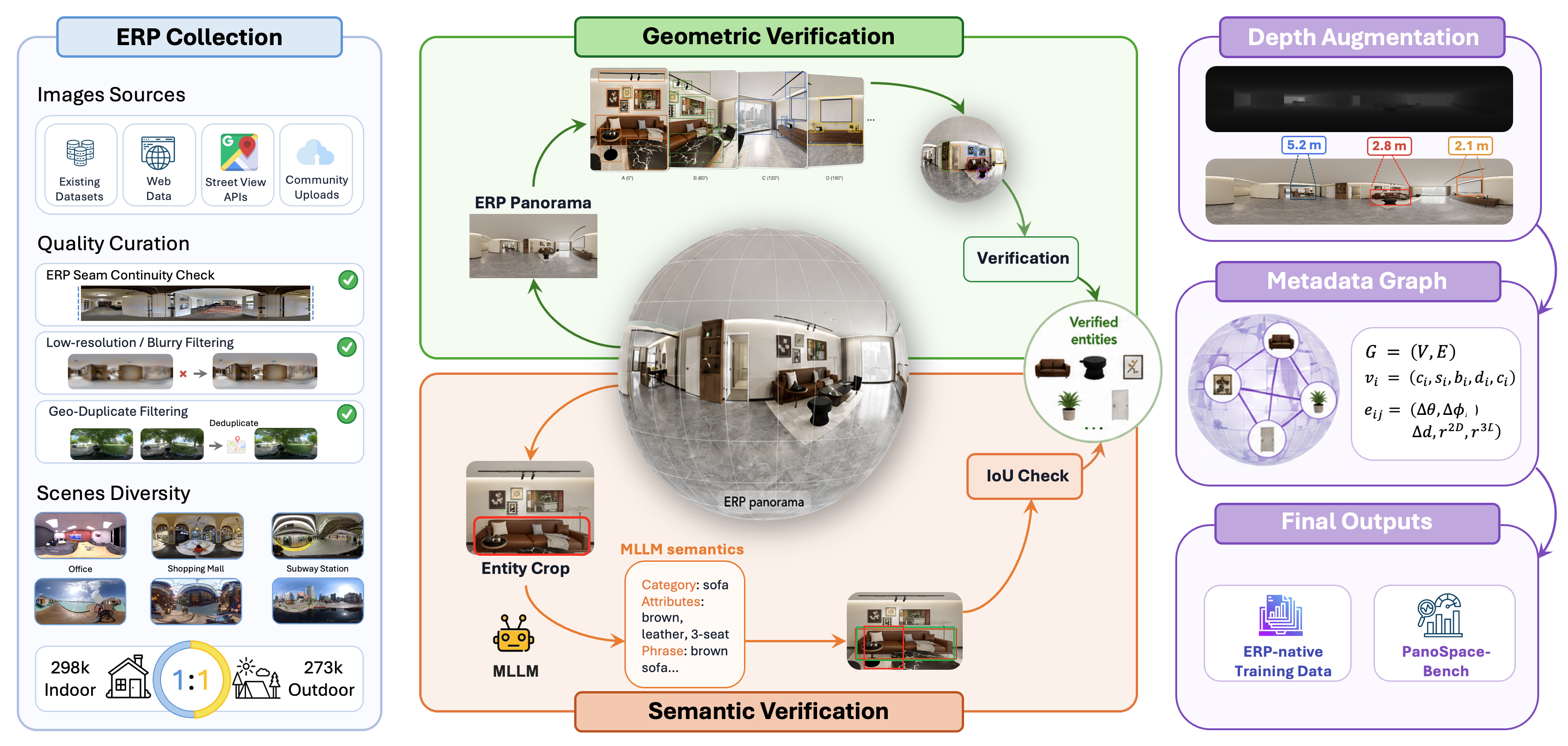} 
    \caption{
    Verifiable metadata construction pipeline. 
    We collect mixed-source ERP panoramas, perform perspective-view detection followed by ERP reprojection and cross-view geometric verification, and enrich verified entities with semantic metadata through MLLM annotation and description-guided referring re-detection. 
    Depth cues are then associated with each entity to build a structured metadata graph, from which both training data and PanoSpace-Bench are derived.
    }
    \label{fig:dataset_pipeline}
\end{figure*}

\subsection{Preliminary and Task Settings}
\label{3.1}

Unlike perspective images defined on a planar image grid, panoramic images are commonly represented in equirectangular projection (ERP), where each pixel corresponds to a spherical direction parameterized by yaw and pitch. For an ERP pixel $(u, v)$ with width $W$ and height $H$, its yaw and pitch are
\begin{equation}
\lambda = 2\pi\left(\frac{u}{W} - \frac{1}{2}\right), \qquad
\phi = \pi\left(\frac{1}{2} - \frac{v}{H}\right),
\end{equation}
where $\lambda \in [-\pi, \pi)$ and $\phi \in [-\frac{\pi}{2}, \frac{\pi}{2}]$. The corresponding unit ray on the sphere is
\begin{equation}
\mathbf{r}(\lambda,\phi) =
\begin{bmatrix}
\cos\phi \sin\lambda \\
\sin\phi \\
\cos\phi \cos\lambda
\end{bmatrix},
\end{equation}
which gives the viewing direction of that ERP location. 

Given an ERP panorama $I \in \mathbb{R}^{H \times W \times 3}$ and a text query $q$, we study pano-native understanding for multimodal large language models by learning a multimodal function
\begin{equation}
    y = f_{\theta}(I, q),
\end{equation}
where $y$ may denote an answer, a direction, a spatial relation, or a grounded target. Different from standard visual question answering on perspective images, this setting requires the model to reason over $I$ as a continuous observer-centered spherical space, including seam continuity, viewpoint reorientation, and relations among entities distributed across the full panorama.

\subsection{Capability Taxonomy for Pano-native Understanding}
\label{3.2}

We decompose pano-native understanding into four capability families that together define the core requirements for reasoning over ERP panoramas. This taxonomy serves as the foundation for both supervision design and benchmark construction.

\textbf{Semantic anchoring.}
The model must ground language to visual entities in ERP panoramas, covering object identity, attributes, scene contents, and global scene-topology semantics such as environment and layout structure. This forms the semantic basis for subsequent spatial reasoning.

\textbf{Spherical grounding.}
The model must localize entities on the observer-centered viewing sphere, where directions are parameterized by yaw and pitch, rather than only on a planar image grid. This ranges from coarse directional localization to fine-grained BFOV-style angular grounding.

\textbf{Reference-frame transformation.}
The model must reason about how spatial relations change under observer rotation or object-conditioned reorientation, including angular relations on the sphere and seam-aware wrap-around continuity.

\textbf{Depth-aware 3D spatial reasoning.}
The model must connect spherical observations to surrounding 3D structure, including depth, relative distance, and viewer-centered relations such as left/right, front/behind, and above/below.

Together, these four families define pano-native understanding from \emph{what} is present, to \emph{where} it lies on the sphere, to \emph{how} its relation changes under reference-frame transformation, and finally to \emph{how} it is organized in 3D space around the observer. Table~\ref{tab:supervision_operators} summarizes the resulting task operators under each family, which instantiate this taxonomy as structured supervision. The next subsection describes how we construct the verified panorama metadata that supports them.

% \vspace{-1em} 

\subsection{Large-scale Dataset Collection and Verifiable Metadata Construction}
\label{3.3}
To support the capability taxonomy above, we require supervision that is both large-scale and verifiable. As shown in Figure~\ref{fig:dataset_pipeline}, we construct a large ERP corpus and derive from it geometry-aware, semantic, and depth-aware metadata, which are finally unified as a structured metadata graph. Please refer to Appendix \ref{Dataset_details} for more details.

\textbf{ERP collection from mixed sources.}
We build a large-scale ERP corpus from mixed sources, including existing panoramic datasets, web data, street-view APIs, and community-contributed uploads, as illustrated in Figure~\ref{fig:dataset_pipeline}. The source composition and scene breakdown are summarized in Table~\ref{tab:erp_corpus_composition}.
We then apply a quality-curation stage to remove invalid or low-quality samples, including ERP seam discontinuity checking, low-resolution and blur filtering, and geo-duplicate removal. Finally, we promote scene diversity by balancing indoor and outdoor panoramas and covering a broad range of environments, such as offices, shopping malls, subway stations, streets, public spaces, and natural scenes. The resulting corpus contains about 570K high-quality ERP panoramas with an approximately balanced indoor/outdoor ratio.

\textbf{Geometry-aware detection metadata.}
Direct detection on ERP is unreliable as object shapes are distorted near high latitudes and may be split by the left-right seam. We therefore project each panorama into a set of overlapping perspective views and apply an off-the-shelf open-vocabulary detector to obtain candidate boxes. The detections are then reprojected to the ERP coordinate system and merged across views. As shown in Figure~\ref{fig:dataset_pipeline}, we further apply \textbf{geometric verification}, including confidence thresholding, IoU-based duplicate suppression, and cross-view consistency checking, to remove unstable proposals caused by projection artifacts, seam splitting, or single-view detector failures. This process produces panorama-level entity candidates with reliable spherical locations and box extents.

\textbf{Language-grounded semantic metadata.}
For each retained candidate, we select the most informative local crop or perspective view and prompt a multimodal language model to generate semantic annotations, including object category, attributes, descriptions, and a discriminative referring phrase. 
We then perform a crop-centered description--re-detection \textbf{semantic verification} step as shown in Figure~\ref{fig:dataset_pipeline}, in which the generated phrase is fed to a referring/open-vocabulary detector to localize the same target again. Candidates whose re-detected boxes do not sufficiently overlap with the original proposals are discarded. This step improves semantic precision and filters out mismatches between language and detection.

\textbf{Depth-aware spatial metadata.}
We further associate each verified entity with depth information. When aligned depth is available from the source data, we use it directly; otherwise, we estimate pseudo-depth with a panoramic depth model \cite{lin2025depth}. Depth values are aggregated over the ERP support region of each entity to estimate observer distance and derive depth-aware spatial cues.

\textbf{Metadata graph construction.}
Combining semantics, angular location, box extent, and depth, we represent each panorama as a structured metadata graph
\begin{equation}
\mathcal{G} = (\mathcal{V}, \mathcal{E}),
\end{equation}
where each node $v_i \in \mathcal{V}$ is a verified entity
\begin{equation}
v_i = \left(s_i, a_i, b_i, d_i, c_i\right),
\end{equation}
with semantics $s_i$, attributes $a_i$, angular footprint $b_i=(\theta_i,\phi_i,\Delta\theta_i,\Delta\phi_i)$, observer distance $d_i$, and local visual context $c_i$.
Each edge $e_{ij}\in\mathcal{E}$ stores pairwise spherical and 3D relations:
\begin{equation}
e_{ij} = \left(\Delta\theta_{ij}, \Delta\phi_{ij}, \Delta d_{ij}, r^{2D}_{ij}, r^{3D}_{ij}\right).
\end{equation}
where $\Delta\theta_{ij}$ and $\Delta\phi_{ij}$ are spherical angular offsets, $\Delta d_{ij}$ is the relative depth difference, and $r^{2D}_{ij}$ and $r^{3D}_{ij}$ denote discretized spherical and viewer-centered 3D relations, respectively. All downstream training tasks are instantiated from this graph, which serves as the structured interface between raw ERP data and capability-aligned supervision. We next describe the pano-aware model adaptation that learns from this supervision.

\begin{figure*}[t]
    \centering
    \includegraphics[width=0.98\textwidth]{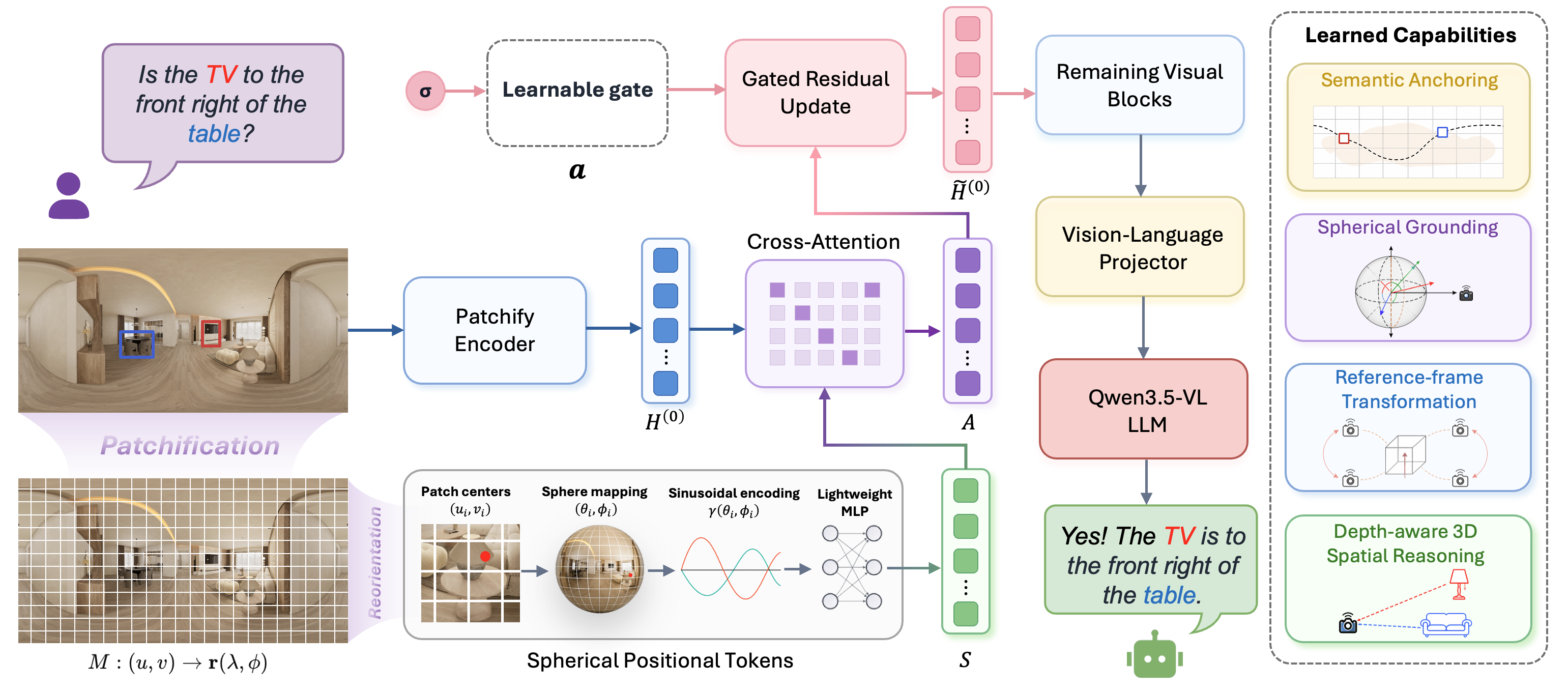} 
    \caption{
    The architecture of PanoWorld. 
    After patch embedding, visual tokens $H^{(0)}$ query spherical spatial tokens $S$ derived from ERP patch centers, producing a geometry-aware signal that is fused through a gated residual update. 
    The enhanced tokens are then fed into the remaining pretrained visual encoder, enabling pano-aware spatial reasoning while preserving the original backbone.
    }
    \label{fig:architecture}
\end{figure*}
% \vspace{1em} 

% 这部分增加了一块即插即用的球面位置编码模块，不是单纯给 patch 加一个 spherical positional embedding，而是提出一个 plug-in 的 Spherical Spatial Cross-Attention (SSCA)，作为 Qwen3.5 视觉流中的几何侧通道。它让视觉 patch token 在进入深层视觉编码前，选择性地读取球面空间 token，从而在不破坏原有 MLLM 视觉语言能力的情况下，引入 ERP 的 observer-centered geometry.

\subsection{Pano-aware MLLM Adaptation}
\label{3.4}

% The taxonomy in Sec.~\ref{3.2} defines \emph{what} should be learned, and the pipeline in Sec.~\ref{3.3} provides \emph{from where} this supervision is obtained. We now describe how the MLLM is adapted to absorb these pano-native signals.

The model are illustrated in Figure \ref{fig:architecture}. We adopt Qwen3.5-VL as the backbone and extend it with a pano-aware module that injects spherical geometry into the visual stream. Since the native visual encoder operates on a planar raster, it does not explicitly account for the spherical structure of ERP images, where the same pixel displacement may correspond to different angular changes at different latitudes and the left and right image borders are adjacent in the real scene. To address this mismatch, we introduce \textbf{Spherical Spatial Cross-Attention (SSCA)}, a pano-aware adapter inserted immediately after patch embedding.

\textbf{Spherical spatial token construction.}
Given an ERP panorama, let $H^{(0)}\in\mathbb{R}^{N\times d}$ denote the patch embeddings produced by the visual patch projector, where $N$ is the number of visual patches and $d$ is the hidden dimension. For each patch $i$, we compute its center $(u_i,v_i)$ in ERP image coordinates and map it to the corresponding spherical direction $(\lambda_i,\phi_i)$. We then encode this direction using a fixed sinusoidal spherical encoding $\gamma(\cdot)$ and project it into the visual hidden space:
\begin{equation}
    s_i = \mathrm{MLP}\!\left(\gamma(\lambda_i,\phi_i)\right) \in \mathbb{R}^{d}.
\end{equation}
Stacking all patch-level spherical tokens gives
\begin{equation}
    S=[s_1,\ldots,s_N]\in\mathbb{R}^{N\times d}.
\end{equation}
Unlike standard 2D positional indices, these tokens are explicitly tied to directions on the viewing sphere. They therefore provide the model with observer-centered geometric cues that remain aligned with the ERP representation.

\textbf{Cross-attention fusion after patch embedding.}
SSCA injects spherical geometry by allowing visual tokens to retrieve information from the spherical tokens through cross-attention:
\begin{equation}
    A = \mathrm{MHA}\big(
    Q=\mathrm{LN}(H^{(0)}),
    K=\mathrm{LN}(S),
    V=\mathrm{LN}(S)
    \big).
\end{equation}
The resulting geometry-aware signal is fused back into the visual stream through a gated residual update:
\begin{equation}
    \widetilde{H}^{(0)} = H^{(0)} + \boldsymbol{\alpha}\odot A,
\end{equation}
where $\boldsymbol{\alpha}\in\mathbb{R}^{d}$ is a learnable gate initialized with a small value.
The updated tokens $\widetilde{H}^{(0)}$ are then fed into the remaining visual blocks. In this way, spherical geometry is injected into the visual stream through adaptive interaction between visual content and observer-centered spatial tokens, while the pretrained backbone remains unchanged. The adapted model is trained on the pano-native instruction corpus derived from Sec.~\ref{3.2} and Sec.~\ref{3.3}.

\begin{table*}[t]
\centering
\small
\setlength{\tabcolsep}{3.0pt}
\renewcommand{\arraystretch}{1.08}
\caption{
Quantitative comparison on PanoSpace-Bench. 
We report overall multiple-choice accuracy, BFOV mean IoU, and category-wise performance across panoramic localization, spherical relational reasoning, omnidirectional 3D spatial reasoning, and ERP representation properties.
}
\resizebox{\textwidth}{!}{
\begin{tabular}{l c cc cccc ccc c}
\toprule
\multirow{2}{*}{Method}
& \multirow{2}{*}{Overall}
& \multicolumn{2}{c}{Localization}
& \multicolumn{4}{c}{Spherical Relation}
& \multicolumn{3}{c}{3D Spatial}
& ERP \\
\cmidrule(lr){3-4}
\cmidrule(lr){5-8}
\cmidrule(lr){9-11}
\cmidrule(lr){12-12}
& 
& Abs. Dir. & BFOV
& Rel. Dir. & Rot. & Reori. & Avg.
& Dist. & Rel. 3D & Avg.
& Seam \\
\midrule
% \multicolumn{12}{l} \\
% \midrule
GPT-4o~\citep{hurst2024gpt4o}               & 31.8 & 37.2 & \underline{17.7} & 34.8 & 28.4 & 24.8 & 29.3 & 45.6 & 27.2 & 36.4 & 37.6 \\
Mimo-v2.5~\citep{mimov25}      & \underline{37.2} & 26.8 & 0.7 & \underline{42.0} & \underline{42.8} & \underline{42.0} & \underline{42.2} & 51.6 & 23.6 & 37.6 & 45.6 \\
% Gemini-3-Pro      & 52.45 & 76.40 & 6.82 & \textbf{60.40} & \textbf{69.60} & \textbf{60.80} & \textbf{63.60} & 53.20 & 31.20 & 42.20 & \textbf{67.60} \\
% \midrule
InternVL3.5-8B~\citep{wang2025internvl35advancingopensourcemultimodal}         & 28.3 & 24.8 & 2.9 & 28.8 & 26.4 & 25.2 & 26.8 & 52.4 & 26.8 & 39.6 & 25.6 \\
% InternVL3.5-8B         &  &  &  &  &  &  &  &  &  &  &  \\
InternVL3.5-14B~\citep{wang2025internvl35advancingopensourcemultimodal}         & 30.4 & 23.6 & 2.8 & 35.6 & 30.8 & 38.4 & 34.9 & \underline{56.8} & 20.8 & 38.8 & 28.8 \\
Qwen2.5-VL-3B~\citep{bai2025qwen25vltechnicalreport}
          & 30.1 & 20.4 & 2.7 & 30.8 & 29.6 & 34.8 & 31.7 & 43.6 & \underline{35.2} & 39.4 & \underline{46.4} \\
Qwen2.5-VL-7B~\citep{bai2025qwen25vltechnicalreport}          & 29.9 & 34.0 & 3.1 & 30.8 & 23.6 & 30.4 & 28.2 & 48.8 & 29.6 & 39.2 & 32.0 \\
Qwen3-VL-8B~\citep{bai2025qwen3vltechnicalreport}            & 29.6 & 47.6 & 2.2 & 28.4 & 21.2 & 27.2 & 25.6 & 45.6 & 23.6 & 34.6 & 34.4 \\
Qwen3-VL-32B~\citep{bai2025qwen3vltechnicalreport}            & 34.8 & 32.0 & 2.5 & 36.4 & 27.2 & 34.8 & 32.8 & 54.8 & 28.8 & \underline{41.8} & 43.6 \\
% Gemma-3             &  &  &  &  &  &  &  &  &  &  &  \\
Ministral-8B~\citep{liu2026ministral3}             & 25.5 & 22.4 & 2.1 & 22.0 & 20.4 & 20.0 & 20.8 & 47.6 & 29.6 & 38.6 & 28.4 \\
Ministral-14B~\citep{liu2026ministral3}             & 23.6 & 35.6 & 2.3 & 20.0 & 14.4 & 12.8 & 15.7 & 49.6 & 31.6 & 40.6 & 21.6 \\
% \midrule
% \multicolumn{12}{l}{\textit{Prompt-enhanced baselines}} \\
\midrule
Qwen3.5-9B~\citep{qwen3.5}
             & 30.8 & 25.2 & 1.4 & 32.2 & 22.6 & 26.3 & 26.1 & 48.6 & 24.5 & 36.9 & 41.2  \\
% + text prompt        &  &  &  &  &  &  &  &  &  &  &  \\
+ visual prompt    & 36.4 & \underline{55.2} & 4.9 & 34.0 & 36.4 & 28.8 & 33.1 & 46.0 & 26.2 & 36.1 & 46.5 \\
% \midrule
% \multicolumn{12}{l}{\textit{ERP-native training}} \\
% \midrule
% Qwen3.5 + naive SFT              &  &  &  &  &  &  &  &  &  &  &  \\
PanoWorld                 & \textbf{56.5}  & \textbf{93.7}  & \textbf{73.3} & \textbf{42.6} & \textbf{52.4} & \textbf{47.2} & \textbf{47.4} & \textbf{59.6} & \textbf{40.6} & \textbf{49.8} &  \textbf{65.5} \\
\bottomrule
\end{tabular}}
\label{tab:main_benchmark_results}
\end{table*}
% \vspace{-1em}

\vspace{-0.5em}
\section{Experiments}
\vspace{-0.5em}

\subsection{Experimental Setup}
\vspace{-0.5em}
\textbf{Training setup.}
Unless otherwise specified, we adopt Qwen3.5 as the base model and fine-tune it on the pano-native instruction corpus constructed in Sec.~\ref{3.3}. All model variants use the same training data mixture and optimization setting for fair comparison. We train on 8 A100 GPUs with AdamW, a learning rate of $1\times10^{-6}$, global batch size 2, gradient accumulation 4, and 1 training epoch.

\textbf{Evaluation benchmarks and metrics.}
We evaluate on three benchmarks: the proposed PanoSpace-Bench, H$^\ast$Bench~\cite{yu2025thinking360}, and R2R-CE Val-Unseen~\cite{krantz2020navgraphvisionandlanguagenavigationcontinuous}.
PanoSpace-Bench covers panoramic localization, spherical relational reasoning, omnidirectional 3D spatial reasoning, and ERP representation properties.
Please refer to Sec.~\ref{app:benchmark_setting} in the Appendix for some details.
We report category-wise accuracy for multiple-choice tasks and BFOV mIoU for fine-grained grounding.
On H$^\ast$Bench, we follow the official protocol and report overall accuracy together with the HOS and HPS subsets.
For R2R-CE, we evaluate VLN transfer using standard navigation metrics, including NE, OSR, SR, and SPL. More details are provided in Appendix~\ref{app:benchmark_setting}.

\vspace{-0.3em}
\subsection{Experimental Results}
\vspace{-0.3em}

\textbf{Quantitative comparison on PanoSpace-Bench.}
We first compare against proprietary MLLMs, open-source MLLMs, prompt-enhanced baselines, and our pano-native model on PanoSpace-Bench.

Table~\ref{tab:main_benchmark_results} shows that general-purpose MLLMs remain weak on pano-native spatial reasoning. Across both proprietary and open-source models, performance drops most clearly on BFOV grounding, reference-frame transformation, and viewer-centered 3D reasoning, even when basic object recognition is relatively strong. This gap indicates that the main challenge is not object semantics alone, but reasoning over the ERP panorama as a continuous observer-centered representation.

Prompt enhancement improves direct ERP inference, especially for coarse localization, confirming that part of the difficulty lies in the missing spherical coordinate convention. However, these gains remain limited on spherical relational reasoning and 3D spatial reasoning, where simply describing the ERP layout is insufficient. See Appendix~\ref{prompt} for visual prompt.

Our pano-native model achieves the best overall performance, improving the Qwen3.5 baseline from 30.8 to 56.5. The gains are broad rather than category-specific: absolute direction rises from 25.2 to 93.7, BFOV mIoU from 1.41 to 73.3, spherical relation average from 26.1 to 47.4, 3D spatial average from 36.9 to 49.8, and seam reasoning from 41.2 to 65.5. These results support the central claim of the paper: robust panoramic understanding requires pano-native spatial learning rather than treating ERP as a wide 2D image or relying only on prompt-level coordinate descriptions.

\begin{table*}[t]
\centering
\tiny
\setlength{\tabcolsep}{2.6pt}
\renewcommand{\arraystretch}{0.84}
\caption{
Transfer evaluation on H$^\ast$Bench.
\textbf{Left:} Perspective-view baselines and our zero-shot ERP model.
\textbf{Right:} ERP-input baselines, prompt-only variants, and training-based adaptation.
}
\vspace{-0.8em}
\label{tab:thinking360_transfer}

\begin{subtable}[t]{0.44\textwidth}
\centering
\caption{Perspective-based methods}
\vspace{-0.5em}
\resizebox{\linewidth}{!}{
\begin{tabular}{lccc}
\toprule
Method & Overall & HOS & HPS \\
\midrule
GPT-4o~\citep{hurst2024gpt4o}              & 21.3 & 19.7 & 23.6 \\
Gemini-2.5-Pro~\citep{comanici2025gemini25}      & 32.3 & 31.9 & \underline{33.0} \\
InternVL3.5-4B~\citep{wang2025internvl35advancingopensourcemultimodal}      & 3.8  & 3.2  & 4.8  \\
InternVL3.5-8B~\citep{wang2025internvl35advancingopensourcemultimodal}      & 6.7  & 6.4  & 7.2  \\
Qwen3-VL-8B~\citep{bai2025qwen3vltechnicalreport}         & 19.1 & 23.6 & 12.2 \\
Qwen3.5-9B~\citep{qwen3.5}          & 18.9 & 21.8 & 14.5 \\
HVS-3B$^\ast$~\citep{yu2025thinking360}       & \underline{38.4} & \underline{47.3} & 24.9 \\
\midrule
\textbf{Ours zero-shot} 
& \textbf{56.1} & \textbf{61.8} & \textbf{47.5} \\
\bottomrule
\end{tabular}
}
\label{tab:thinking360_main}
\end{subtable}
\hfill
\begin{subtable}[t]{0.53\textwidth}
\centering
\caption{ERP panorama-based methods}
\vspace{-0.5em}
\resizebox{\linewidth}{!}{
\begin{tabular}{lccccc}
\toprule
Route / Setting & Overall & HOS & HPS & Yaw Acc & Pitch Acc \\
\midrule
GPT-4o~\citep{hurst2024gpt4o}             & 30.1 & 39.1 & 17.1 & 38.5 & 64.2 \\
Gemini-2.5-Pro~\citep{comanici2025gemini25}     & \underline{46.9} & \underline{55.3} & \underline{34.3} & \underline{52.5} & \underline{71.6} \\
InternVL3.5-4B~\citep{wang2025internvl35advancingopensourcemultimodal}     & 11.6   & 12.8   & 9.75   & 22.1   & 41.5 \\
InternVL3.5-8B~\citep{wang2025internvl35advancingopensourcemultimodal}     & 14.9   & 18.0   & 10.2   & 19.2   & 38.5 \\
Qwen3-VL-8B~\citep{bai2025qwen3vltechnicalreport}        & 13.1   & 15.0   & 10.3   & 16.8   & 39.2 \\
\midrule
Qwen3.5-9B~\citep{qwen3.5}         & 19.4 & 26.2 & 9.3  & 23.5 & 46.5 \\
+ Text prompt      & 38.5 & 43.3 & 31.2 & 40.0 & 49.5 \\
+ Visual prompt    & 40.4 & 46.0 & 32.0 & 43.5 & 52.0 \\
\midrule
Qwen3.5 + H$^\ast$ SFT 
                   & 17.8 & 11.1 & 27.7 & 25.3 & 42.5 \\
\textbf{Ours + H$^\ast$ SFT}
& \textbf{70.1} & \textbf{73.1} & \textbf{64.2} & \textbf{74.1} & \textbf{85.5} \\
\bottomrule
\end{tabular}
}
\label{tab:thinking360_routes}
\end{subtable}
\vspace{-2em}
\end{table*}

\begin{wrapfigure}{r}{0.48\columnwidth}
    \vspace{-0.8em}
    \centering
    \includegraphics[width=0.48\columnwidth]{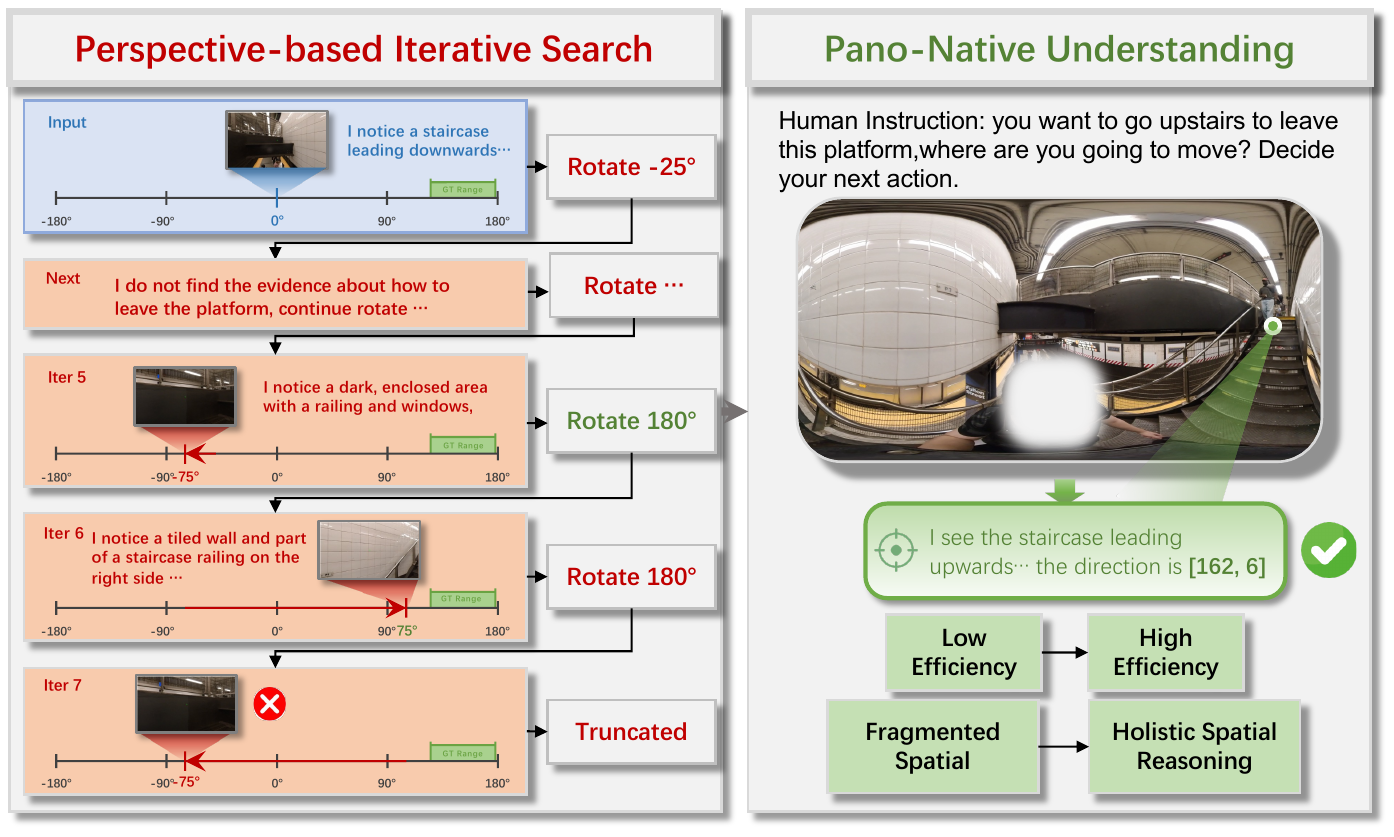}
    \caption{
    Case comparison on H$^\ast$Bench. 
    Perspective-view iterative search is inefficient and may fail due to fragmented local observations, whereas direct ERP input enables holistic reasoning and correct prediction in one step.
    }
    \label{fig:hstar_case}
    \vspace{-1.0em}
\end{wrapfigure}

\textbf{Quantitative comparison on H$^\ast$Bench.}
We evaluate transfer to H$^\ast$Bench~\cite{yu2025thinking360}, which contains Humanoid Object Search (HOS) and Humanoid Path Search (HPS). As shown in Figure~\ref{fig:hstar_case}, prior methods are typically evaluated through perspective-view exploration, whereas our model directly takes ERP panoramas as input. This tests whether the learned representation transfers beyond our own benchmark and supports downstream 360$^\circ$ human-centric visual search.

Table~\ref{tab:thinking360_transfer}(a) compares the conventional perspective-view protocol with direct ERP input. Our zero-shot model reaches 56.10 overall, substantially outperforming the strongest reported perspective-view baseline (38.40). 
% After H$^\ast$Bench fine-tuning, performance further rises to 70.00 overall, 73.00 on HOS, and 64.00 on HPS. 
This shows that the representation learned from pano-native supervision transfers to downstream panoramic search rather than overfitting to PanoSpace-Bench alone.

Table~\ref{tab:thinking360_transfer}(b) shows that direct ERP inference with a generic Qwen3.5 model reaches only 19.4 overall, and prompt engineering improves it to 38.5--40.4, suggesting that explicit coordinate instructions help but do not solve pano-native reasoning. Naive H$^\ast$Bench fine-tuning with ERP input performs even worse (17.8 overall), indicating that target-task supervision alone is insufficient when the base model lacks panoramic spatial priors. In contrast, our model reaches 56.10 zero-shot and 70.00 after H$^\ast$ fine-tuning. Together, these results show that pano-native learning provides a transferable spatial initialization that is not recoverable from prompting or naive task-specific supervision.

\textbf{Quantitative comparison on VLN.}
We further evaluate transfer on R2R-CE Val-Unseen using only ERP panorama as input, which is different from methods that use panoramas to sample candidate perspective views for viewpoint selection.
\begin{table}[t]
\centering
\scriptsize
\setlength{\tabcolsep}{2.6pt}
\renewcommand{\arraystretch}{0.98}
\caption{
R2R-CE Val-Unseen evaluation.
$\dagger$ indicates waypoint-predictor methods; 
$*$ indicates training only on R2R/RxR for fair comparison.
We follow this setting and use only 80\% training data.
}
\label{tab:vln_r2r_ce}
\begin{tabular*}{\columnwidth}{@{\extracolsep{\fill}} l c c c c c c c c c c @{}}
\toprule
\multirow{2}{*}{\textbf{Method}} 
& \multirow{2}{*}{\textbf{Year}}
& \multirow{2}{*}{\textbf{Paradigm}}
& \multicolumn{4}{c}{\textbf{Observation}} 
& \multicolumn{4}{c}{\textbf{R2R-CE Val-Unseen}} \\
\cmidrule(lr){4-7} \cmidrule(lr){8-11}
& & 
& \textbf{Pano.} & \textbf{Odom.} & \textbf{Depth} & \textbf{RGB}
& \textbf{NE}$\downarrow$ & \textbf{OSR}$\uparrow$ & \textbf{SR}$\uparrow$ & \textbf{SPL}$\uparrow$ \\
\midrule

HPN+DN$^\dagger$~\citep{krantz2021waypoint}
& 2021 & Waypoint
& \cmark & \cmark & \cmark & 
& 6.31 & 40.0 & 36.0 & 34.0 \\

Sim2Sim$^\dagger$~\citep{krantz2022sim2sim}
& 2022 & Waypoint
& \cmark & \cmark & \cmark & 
& 6.07 & 52.0 & 43.0 & 36.0 \\

GridMM$^\dagger$~\citep{wang2023gridmm}
& 2023 & Waypoint
& \cmark & \cmark & \cmark & 
& 5.11 & 61.0 & 49.0 & 41.0 \\

DreamWalker$^\dagger$~\citep{wang2023dreamwalker}
& 2023 & Waypoint
& \cmark & \cmark & \cmark & 
& 5.53 & 59.0 & 49.0 & 44.0 \\

\midrule

Uni-NaVid$^*$~\citep{zhang2024uninavid}
& 2024 & RGB-only
&  &  &  & \cmark
& 5.58 & 53.3 & 47.0 & 42.7 \\

StreamVLN$^*$~\citep{wei2025streamvln}
& 2025 & RGB-only
&  &  &  & \cmark
& 5.73 & 56.4 & 50.2 & 47.1 \\

Efficient-VLN$^*$~\citep{zheng2025efficientvln}
& 2026 & RGB-only
&  &  &  & \cmark
& 6.41 & 54.5 & 45.9 & 41.9 \\

NaVIDA$^*$~\citep{navida2026}
& 2026 & RGB-only
&  &  &  & \cmark
& 5.72 & 57.4 & 47.7 & 41.5 \\

\midrule
\textbf{PanoWorld-VLN$^*$}
& 2026 & \textbf{Direct ERP}
& \cmark &  &  & 
& \textbf{4.98} & \textbf{59.3} & \textbf{54.3} & \textbf{52.1} \\
\bottomrule
\end{tabular*}
\end{table}
As shown in Table~\ref{tab:vln_r2r_ce}, PanoWorld achieves 54.3 SR and 52.1 SPL using only panorama input. 
Compared with methods that rely on waypoint predictors or use panoramas mainly for candidate-view selection, PanoWorld directly consumes the ERP panorama as a unified full-surround observation and improves over GridMM by 5.3 SR and 11.1 SPL. 
It also outperforms recent RGB-only VLN models under the same R2R/RxR training setting, surpassing StreamVLN by 4.1 SR and 5.0 SPL despite using only 80\% of the training data. 
These results suggest that direct panoramic understanding provides an efficient and transferable paradigm for VLN, moving beyond fragmented local-view exploration toward unified full-surround spatial reasoning.

\vspace{-0.5em}
\subsection{Ablation Studies}
\vspace{-0.5em}

We ablate the proposed framework from four perspectives: the composition of ability-oriented training data, the verification modules in metadata construction, the architecture used to inject spherical geometry, and the trainable scope used for ERP adaptation. Unless otherwise specified, all ablations are evaluated on the same split of PanoSpace-Bench. 
Detailed ablations on metadata verification, pano-aware architecture, and trainable scope are provided in Appendix~\ref{sec:appendix_ablation}.

% 这一节核心claim的点:证明我们提出的这四个基本能力是必要的，内容还需优化：

\begin{table*}[t]
\centering
\small
\setlength{\tabcolsep}{4.0pt}
\renewcommand{\arraystretch}{1.08}
\caption{
Ability-oriented training data ablation.
The left block shows the four training ability groups derived from our pano-native capability taxonomy.
}
\resizebox{\textwidth}{!}{
\begin{tabular}{c c c c | c c | c c c c | c c c | c}
\toprule
\multicolumn{4}{c|}{Training Data Ability} 
& \multicolumn{2}{c|}{Localization} 
& \multicolumn{4}{c|}{Spherical Relation} 
& \multicolumn{3}{c|}{3D Spatial} 
& ERP \\
\cmidrule(lr){1-4}
\cmidrule(lr){5-6}
\cmidrule(lr){7-10}
\cmidrule(lr){11-13}
\cmidrule(lr){14-14}
Semantic 
& Spherical 
& Ref.-Frame 
& 3D Spatial 
& Abs. Dir. 
& BFOV mIoU 
& Rel. Dir. 
& Cam. Rot. 
& Obj. Reori. 
& Avg. 
& Dist. 
& Rel. 3D 
& Avg. 
& Seam \\
\midrule

\cmark &  &  &   
& 24.0 & 5.4 
& 33.6 & 17.6 & 27.6 & 26.3 
& 45.2 & 24.0 & 34.6 
& 43.6 \\ 

 & \cmark &  &   
& 59.2 & 37.6
& 34.8 & 15.6 & 33.2 & 27.9 
& 49.6 & 24.4 & 37.0 
& 48.8 \\

 &  & \cmark &  
& 36.4 & 2.3 
& 38.8 & 29.6 & 38.8 & 35.7 
& 50.4 & 19.6 & 35.0 
& 49.2 \\

 &  &  & \cmark 
% &  &  &  &  &  &  &  &  &  &  \\
& 32.0 & 2.2 & 32.8 & 13.2 & 19.6 & 21.8 & \textbf{63.2} & \textbf{36.8} & \textbf{50.0} & 49.2 \\

\midrule

\cmark & \cmark &  &  
& 67.2 & 50.1 
& 33.2 & 16.0 & 29.6 & 26.3 
& 56.4 & 26.8 & 41.6 
& 44.0 \\

\cmark &  & \cmark &  
& 25.2 & 1.8 & \textbf{41.6} & \underline{38.4} & 41.6 & \textbf{40.5} & \underline{59.2} & 19.6 & 39.4 & 48.8 \\

\cmark &  &  & \cmark 
& 27.6 & 1.1 & 31.2 & 15.6 & 22.0 & 22.9 & 57.2 & 34.0 & 45.6 & 48.8 \\

\midrule

\cmark &  & \cmark & \cmark 
% &  &  &  &  &  &  &  &  &  &  \\
& 36.4 & 5.6 & 37.2 & 29.2 & 40.8 & 35.7 & 53.6 & \underline{35.6} & 44.6 & 47.6 \\

\cmark & \cmark &  & \cmark 
% &  &  &  &  &  &  &  &  &  &  \\
& \textbf{71.6} & \underline{50.7} & 32.0 & 16.0 & 26.8 & 24.9 & 55.6 & 32.0 & \underline{43.8} & 49.2 \\

\cmark & \cmark & \cmark &  
& 60.0 & 43.0 
& \underline{40.0} & 29.2 & \underline{44.8} & 38.0 
& 57.2 & 24.8 & 41.0 
& \underline{51.8} \\ 

\cmark & \cmark & \cmark & \cmark 
& \underline{68.8} & \textbf{66.7} 
& 34.8 & \textbf{49.6} & \textbf{45.2} & \underline{43.2} 
& 52.0 & 34.8 & 43.4 
& \textbf{53.0} \\ 

\bottomrule
\end{tabular}
}
\label{tab:ablation_training_data}
\end{table*}

\textbf{Ability-oriented training data.}
Table~\ref{tab:ablation_training_data} validates the proposed ability decomposition. Semantic-only training yields limited spatial performance, while adding spherical grounding sharply improves localization. Reference-frame transformation data is most beneficial for spherical relational reasoning, and depth-aware 3D supervision improves distance and relative 3D understanding. The best overall results are obtained by combining all ability families, indicating that panoramic understanding is compositional: semantic anchoring alone is insufficient, and strong performance requires jointly learning localization, transformation, and 3D spatial structure.

% 这一节核心claim的点:证明我们的metadata pipeline中的验证模块，可以大大提高数据的质量

\begin{table}[t]
\centering
\small
\caption{
Ablation of verification modules in the metadata construction pipeline.
Detection verification filters geometrically unreliable proposals and grounding targets, while semantic verification removes semantically inconsistent question-answer pairs.
}
\resizebox{\textwidth}{!}{
\begin{tabular}{l c c c c c c c}
\toprule
Pipeline Variant & Det. Verif. & Sem. Verif. & Overall &  Localization & Direction & 3D & Seam \\
\midrule
baseline &  &  & 38.8 & 58.3 & 32.6 & 35.5 & 48.2 \\
w/ Detection Verification & \checkmark &  & 46.4 & \underline{70.5} & 39.9 & 42.9 & 56.1 \\ 
w/ Semantic Verification &  & \checkmark & \underline{48.0} & 70.3 & \underline{41.5} & \underline{43.8} & \underline{58.9} \\
Full pipeline  & \checkmark & \checkmark  & \textbf{55.1} & \textbf{82.7} & \textbf{46.0} & \textbf{49.0} & \textbf{65.5} \\
\bottomrule
\end{tabular}
}
\label{tab:pipeline_ablation}
\end{table}

\textbf{Metadata verification.}
Table~\ref{tab:pipeline_ablation} shows that both verification modules are important for reliable ERP supervision. Starting from the unverified baseline, detection verification improves overall accuracy from 38.8 to 46.4 by filtering geometrically unstable proposals, while semantic verification raises it to 48.0 by removing inconsistent language-region pairs. Combining both yields the best result of 55.1, with gains across localization, directional reasoning, 3D reasoning, and seam continuity. This confirms that data quality is a major factor in pano-native learning.

% 这一节核心claim的点:证明我们的架构设计确实有效，通过球面位置编码的嵌入增强了模型对于全景3d空间的几何理解
% ERP球形几何信息应尽早且自适应地注入。在图像块层面进行交叉注意力，可以让视觉标记在视觉聚合之前查询球形位置线索，这比将几何信息视为静态残余偏差或注入过晚更为有效。

\begin{table}[!t]
\centering
\scriptsize
\setlength{\tabcolsep}{3.2pt}
\renewcommand{\arraystretch}{1.08}
\caption{
Architecture ablation of ERP-aware spherical geometry injection. 
We compare residual addition and cross-attention under three insertion positions: patch, merge, and output. 
}
\resizebox{\textwidth}{!}{
\begin{tabular}{l l c c c c c c c c c c c}
\toprule
Method & Position & Overall 
& \multicolumn{2}{c}{Localization}
& \multicolumn{4}{c}{Spherical Relation}
& \multicolumn{3}{c}{3D Spatial}
& ERP \\
\cmidrule(lr){4-5}
\cmidrule(lr){6-9}
\cmidrule(lr){10-12}
\cmidrule(lr){13-13}
 &  & 
 & Abs. Dir. & BFOV mIoU
 & Rel. Dir. & Cam. Rot. & Obj. Reori. & Avg.
 & Dist. & Rel. 3D & Avg.
 & Seam \\
\midrule
Qwen3.5 
& -- 
& 48.40
& 82.40 & 43.60
& 35.60 & 32.80 & 34.80 & 34.40
& 58.00 & 32.80 & 45.40
& 56.40 \\

\midrule

Residual 
& Merge  
& 48.50
& 68.80 & 38.40
& 36.00 & 35.20 & 45.20 & 38.80
& 57.60 & 32.80 & 45.20
& 64.00 \\

Residual 
& Output 
& 50.50
& 79.20 & 43.10
& \underline{37.60} & 36.40 & 34.80 & 36.30
& \textbf{58.80} & \underline{36.80} & \underline{47.80}
& \textbf{68.00} \\

Residual 
& Patch  
& \underline{51.50}
& \underline{82.40} & 48.10
& \underline{37.60} & 40.00 & 36.00 & 37.90
& 57.20 & 34.00 & 45.60
& 64.00 \\
\midrule

Cross-Attn 
& Merge  
& 50.80
& 81.60 & 43.80
& 34.40 & 42.80 & \textbf{47.60} & 41.60
& 53.20 & 34.40 & 43.80
& 61.60 \\

Cross-Attn 
& Output 
& 51.00
& 81.40
& \underline{45.70}
& 33.00 & \textbf{51.60} & 45.60 & \underline{43.40}
& 52.80 & 34.00 & 43.40
& 58.00 \\

Cross-Attn 
& Patch  
& \textbf{55.10}
& \textbf{92.80} & \underline{72.60}
& \textbf{40.00} & \underline{51.20} & \underline{46.80} & \textbf{46.00}
& \underline{58.40} & \textbf{39.20} & \textbf{48.80}
& \underline{64.80} \\
\bottomrule
\end{tabular}
}
\label{tab:arch_ablation}
\end{table}

% \begin{table}[t]
% \centering
% \small
% \begin{tabular}{l l c c c c c}
% \toprule
% Method & Position & Overall & Localization & Sph. Rel. & 3D & Seam \\
% \midrule
% Vanilla Qwen3.5 & -- 
% & 0.484 & 0.824 & 0.344 & 0.454 & 0.564 \\
% \midrule
% Residual & Patch  
% & 0.515 & 0.916 & 0.379 & 0.456 & 0.640 \\

% Residual & Merge  
% & 0.485 & 0.688 & 0.388 & 0.452 & 0.640 \\

% Residual & Output 
% & 0.517 & 0.892 & 0.363 & 0.478 & \textbf{0.680} \\
% \midrule
% Cross-Attn & Patch  
% & \textbf{0.551} & \textbf{0.928} & \textbf{0.460} & \textbf{0.488} & 0.648 \\

% Cross-Attn & Merge  
% & 0.508 & 0.816 & 0.416 & 0.438 & 0.616 \\

% Cross-Attn & Output 
% &  &  &  &  &  \\
% \bottomrule
% \end{tabular}
% \caption{
% Architecture ablation of ERP-aware spherical geometry injection. 
% We compare residual addition and cross-attention under three insertion positions: patch, merge, and output.
% Overall denotes average multiple-choice accuracy. 
% Localization reports absolute-direction accuracy in this compact table. 
% Sph. Rel. averages relative direction, camera rotation, and object-conditioned reorientation tasks. 
% 3D averages observer-distance and relative 3D-position tasks. 
% Seam reports boundary-continuity accuracy.
% }
% \label{tab:arch_ablation}
% \end{table}

\textbf{Architecture ablation.}
Table~\ref{tab:arch_ablation} compares residual fusion and cross-attention at different insertion positions. Patch-level cross-attention performs best overall, improving accuracy from 0.484 to 0.551 and yielding the strongest spherical relation average (0.460) and 3D spatial average (0.488). Residual fusion also helps in some settings, especially seam continuity, but is less consistent on relation-heavy categories. These results support the proposed SSCA design: geometry is most effective when injected early and through content-dependent interaction with visual tokens.

% 这一节核心claim的点:探究了究竟要训练哪些层才对pano-native learning最有效？
% 

\begin{table}[H]
\centering
\scriptsize
\setlength{\tabcolsep}{3.2pt}
\renewcommand{\arraystretch}{1.05}
\caption{
Trainable scope ablation for pano-native spatial learning.
We report fine-grained performance across benchmark tasks instead of aggregated group averages.
}
\resizebox{\textwidth}{!}{
\begin{tabular}{l c c c | c c | c c c | c c | c}
\toprule
\multirow{2}{*}{Train Scope}
& \multicolumn{3}{c|}{Trainable Components}
& \multicolumn{2}{c|}{Localization}
& \multicolumn{3}{c|}{Spherical Relation}
& \multicolumn{2}{c|}{3D Spatial}
& ERP \\
\cmidrule(lr){2-4}
\cmidrule(lr){5-6}
\cmidrule(lr){7-9}
\cmidrule(lr){10-11}
\cmidrule(lr){12-12}
& Vision & VL Int. & LLM
& Abs. Dir. & BFOV mIoU
& Rel. Dir. & Cam. Rot. & Obj. Reori.
& Dist. & Rel. 3D
& Seam \\
\midrule

LLM only
&  &  & \cmark
% &  &  &  &  &  &  &  &  \\
& \underline{84.80} & 62.28 & 34.00 & 36.40 & 38.00 & \underline{57.60} & \underline{37.20} & 55.60 \\

VL Int. only
&  & \cmark & 
% &  &  &  &  &  &  &  &  \\
& 25.60 & 2.19 & 33.60 & 22.80 & 26.40 & 41.40 & 24.80 & 37.40 \\

VL Int. + LLM
&  & \cmark & \cmark
% &  &  &  &  &  &  &  &  \\
& 83.60 & \underline{64.25} & \underline{35.20} & \underline{47.20} & \underline{44.40} & 56.80 & 35.20 & \underline{56.40} \\

\midrule

Vision + VL Int.
& \cmark & \cmark & 
& 34.20 & 6.04 & 32.40 & 24.80 & 31.30 & 47.60 & 29.20 & 46.40 \\

Full pano-native FT
& \cmark & \cmark & \cmark
& \textbf{92.80} & \textbf{72.60} & \textbf{40.0} & \textbf{51.2} & \textbf{46.8} & \textbf{58.4} & \textbf{39.2} & \textbf{64.8} \\

\bottomrule
\end{tabular}
}
\label{tab:trainable_ablation}
\end{table}

\textbf{Trainable component ablation.} Table~\ref{tab:trainable_ablation} examines the trainable scope during pano-native adaptation. We compare different update strategies over the vision encoder, the vision-language interface, and the language model to identify where ERP-native spatial learning mainly takes place. The results indicate that panoramic reasoning depends not only on language-side adaptation, but also on updating the visual and cross-modal components that encode and align ERP geometry.

\section{Conclusion}

In this paper, we introduce a unified pano-native spatial learning framework for multimodal large language models. We formulate pano-native understanding as reasoning over ERP panoramas as continuous observer-centered spaces, and decompose it into four core capability families: semantic anchoring, spherical grounding, reference-frame transformation, and depth-aware 3D spatial reasoning. Based on this formulation, we build a large-scale metadata construction pipeline, derive capability-aligned instruction-tuning data, and propose an pano-aware model adaptation with spherical spatial cross-attention. We further construct PanoSpace-Bench to evaluate pano-native spatial reasoning beyond conventional VQA-style settings. Extensive experiments show that the proposed framework substantially improves panoramic reasoning on the proposed PanoSpace-Bench, H$^\ast$Bench, and R2R-CE Val-Unseen benchmarks. 

\clearpage

\medskip

\bibliographystyle{plainnat}
\bibliography{related_works_refs}

%%%%%%%%%%%%%%%%%%%%%%%%%%%%%%%%%%%%%%%%%%%%%%%%%%%%%%%%%%%%

\appendix

\clearpage

\section{Dataset Details}
\label{Dataset_details}

\subsection{ERP Corpus Composition}

Table~\ref{tab:erp_corpus_composition} summarizes the ERP image sources. Our ERP corpus contains 570,321 full-surround panoramas collected from mixed sources. 
The corpus is approximately balanced between indoor and outdoor scenes, with 297,476 indoor panoramas and 272,845 outdoor panoramas. 
This balance is important for pano-native spatial learning, since indoor scenes provide object-rich local layouts and depth relations, while outdoor scenes introduce larger-scale structures, long-range visibility, and diverse panoramic topology.

Our work uses large-scale panoramic imagery, which raises licensing, privacy, and misuse considerations.
The ERP corpus combines existing panoramic datasets and public web sources, including Realsee3D~\citep{Li2025realsee3d_data} and 360+X~\citep{chen2024x360}.
We cite the external datasets and model components used in the paper and will release only assets whose redistribution is compatible with the corresponding source licenses, data-use agreements, and terms of service.
Panoramic imagery may contain homes, bystanders, vehicles, or other sensitive visual details.
Before release, we will apply privacy-oriented filtering to remove or mask personally identifying content where applicable, exclude unsafe or sensitive scenes, and provide a removal channel for reported problematic examples.

\vspace{-2em}
% \begin{table}[t]
\begin{table}[H]
\centering
\small
\setlength{\tabcolsep}{4.5pt}
\renewcommand{\arraystretch}{1.08}
\caption{Composition of the collected ERP panorama corpus.}
\label{tab:erp_corpus_composition}
\begin{tabular}{lccc}
\toprule
\textbf{Source} & \textbf{\#Panoramas} & \textbf{Ratio} & \textbf{Scene type} \\
\midrule
Realsee3D-real~\citep{Li2025realsee3d_data}
& 24,025  & 4.2\%  & Indoor \\

Realsee3D-synthetic~\citep{Li2025realsee3d_data}
& 273,451 & 47.9\% & Indoor \\

360+X~\citep{chen2024x360}
& 15,290  & 2.7\%  & Outdoor \\

Outdoor web crawling
& 76,643  & 13.4\% & Outdoor \\

Street-view web crawling
& 63,651  & 11.2\% & Outdoor \\

API-based collection
& 117,261 & 20.6\% & Outdoor \\
\midrule
\textbf{Indoor total} 
& \textbf{297,476} & \textbf{52.2\%} & Indoor \\

\textbf{Outdoor total} 
& \textbf{272,845} & \textbf{47.8\%} & Outdoor \\
\midrule
\textbf{Total} 
& \textbf{570,321} & \textbf{100.0\%} & Mixed \\
\bottomrule
\end{tabular}
\end{table}
\vspace{-2em}

% \begin{figure}[H]
%     \centering
%     \begin{minipage}[t]{0.49\columnwidth}
%         \centering
%         \includegraphics[width=\textwidth]{figure/supp/object_frequency.png}
%         \caption{Object category distribution in the constructed metadata.}
%         \label{fig:object_frequency}
%     \end{minipage}
%     \hfill
%     \begin{minipage}[t]{0.49\columnwidth}
%         \centering
%         \includegraphics[width=\textwidth]{figure/supp/qa_form_distribution.png}
%         \caption{Instruction format distribution in the generated training data.}
%         \label{fig:qa_form_distribution}
%     \end{minipage}
% \end{figure}

\subsection{Metadata Pipeline Details}
\label{metadate_supp}

We provide additional implementation details for the metadata construction pipeline in Sec.~\ref{3.3}. 
For each ERP panorama, we render overlapping perspective views with a $120^\circ$ FoV and a $60^\circ$ yaw stride, resulting in approximately $60^\circ$ overlap between adjacent views. 
We use WeDetect-Large~\citep{fu2025wedetect} as the open-world detector, with a confidence threshold of $0.3$ and a view-level NMS IoU threshold of $0.5$. 
The detected boxes are reprojected to ERP coordinates and merged across overlapping views; two boxes are considered geometrically consistent if their ERP IoU exceeds $0.6$. 
For semantic annotation, we use Qwen3-VL-32B~\citep{bai2025qwen3vltechnicalreport} to generate object categories, attributes, descriptions, and discriminative referring phrases. 
We then use WeDetect-Ref-4B~\citep{fu2025wedetect} for description-guided re-detection and retain an entity only if the IoU between the original proposal and the re-detected box exceeds $0.7$. 
This provides a strong implementation of geometric and language verification before constructing the final metadata graph~\citep{wang2026rationalrewards}.

\subsection{Instruction Data Distribution}

\begin{figure}[H]
    \centering
    \includegraphics[width=\columnwidth]{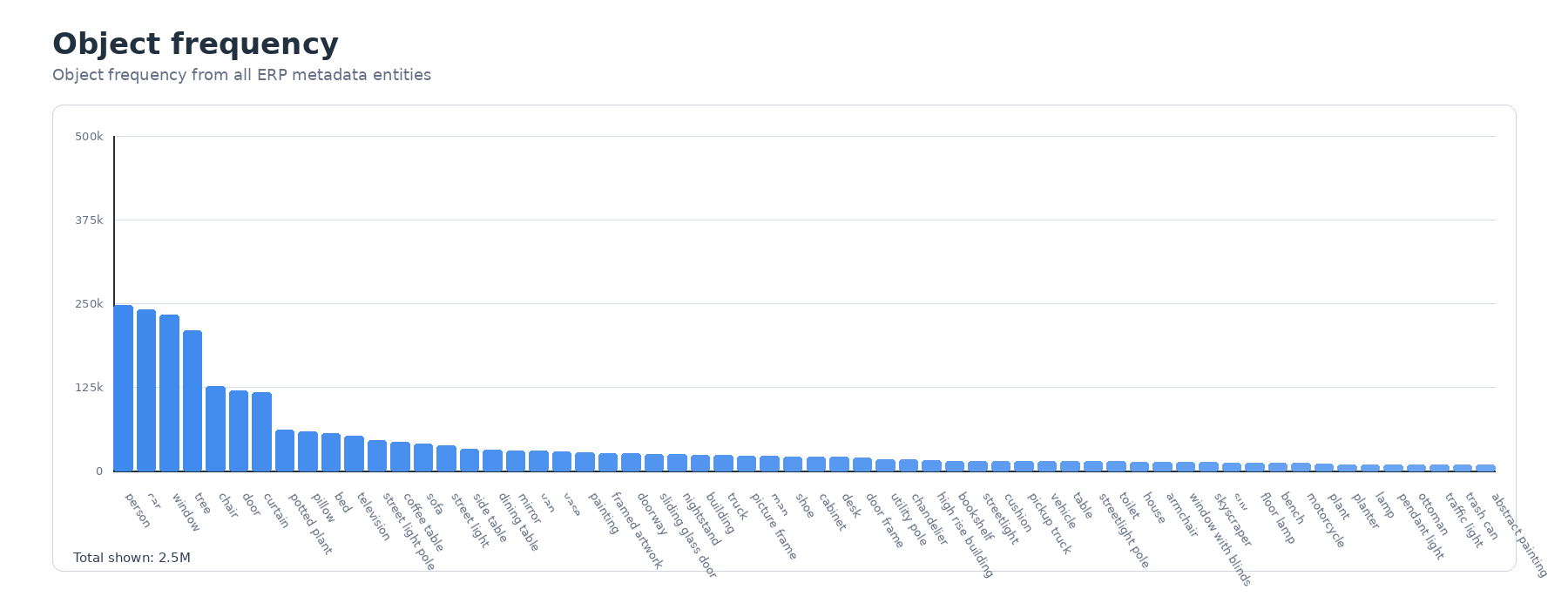}
    \caption{Object category distribution in the constructed metadata.}
    \label{fig:object_frequency}
\end{figure}

\begin{figure}[H]
    \centering
    \includegraphics[width=\columnwidth]{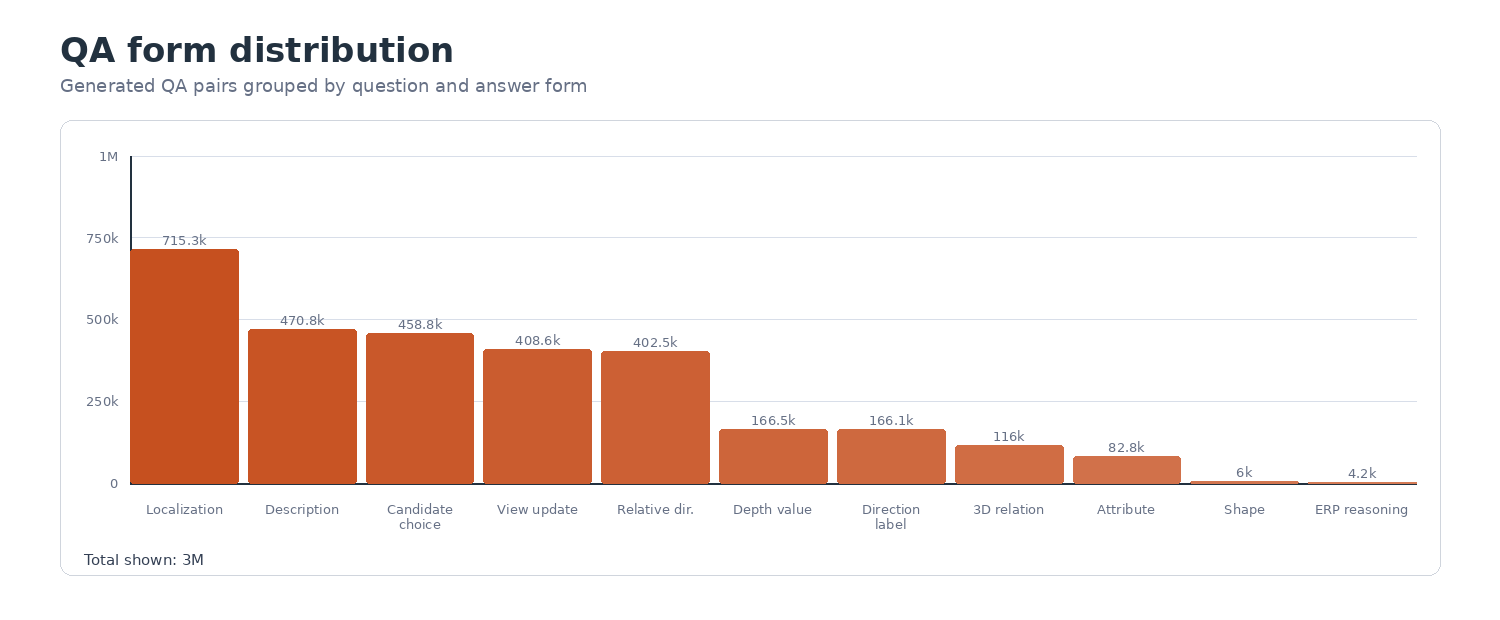}
    \caption{Instruction format distribution in the generated training data.}
    \label{fig:qa_form_distribution}
\end{figure}

Table~\ref{tab:supervision_operators} summarizes the ERP corpus composition, and Table~\ref{tab:instruction_family_distribution} reports the final instruction distribution.
From the verified metadata graphs, we first instantiate 7.65M candidate instruction samples and then sample a canonical training set of 2.998M examples. 
The sampling procedure targets a balanced mixture across semantic, angular, reference-frame, and depth-aware reasoning tasks, while limiting repeated samples from the same scene within each family. 
On average, each ERP panorama contributes 5.26 canonical instruction examples.
Fig.~\ref{fig:object_frequency} and Fig.~\ref{fig:qa_form_distribution} illustrate the frequency of entity categories and the distribution of instruction formats.

\vspace{-2em}

\vspace{-2em}

\begin{table}[t]
\centering
\small
\setlength{\tabcolsep}{4.2pt}
\renewcommand{\arraystretch}{1.08}
\caption{
Distribution of pano-native instruction data by ability family.
We first instantiate a large candidate pool from metadata graphs, and then sample a canonical training set according to the target task mixture.
}
\label{tab:instruction_family_distribution}
\begin{tabular}{lcccc}
\toprule
\multirow{2}{*}{\textbf{Ability family}} 
& \multicolumn{2}{c}{\textbf{Candidate pool}} 
& \multicolumn{2}{c}{\textbf{Canonical set}} \\
\cmidrule(lr){2-3} \cmidrule(lr){4-5}
& \textbf{Examples} & \textbf{Ratio} 
& \textbf{Examples} & \textbf{Ratio} \\
\midrule
Semantic anchoring 
& 2,120,452 & 27.7\% 
& 1,101,882 & 36.8\% \\

Angular grounding 
& 1,060,226 & 13.9\% 
& 333,091 & 11.1\% \\

Spherical reference-frame transformation 
& 2,097,356 & 27.4\% 
& 823,810 & 27.5\% \\

Depth-aware 3D relation 
& 2,352,181 & 30.8\% 
& 732,730 & 24.4\% \\

ERP distortion/topology awareness 
& 15,852 & 0.2\% 
& 6,003 & 0.2\% \\
\midrule
\textbf{Total} 
& \textbf{7,646,067} & \textbf{100.0\%} 
& \textbf{2,997,516} & \textbf{100.0\%} \\
\bottomrule
\end{tabular}
\end{table}

\begin{table}[t]
\centering
\footnotesize
\setlength{\tabcolsep}{3pt}
\renewcommand{\arraystretch}{1.18}
\caption{Detailed task templates instantiated by each pano-native supervision operator.}
\label{tab:supervision_operators}
\begin{tabularx}{\linewidth}{@{}L{0.17\linewidth} L{0.20\linewidth} Y Y@{}}
\toprule
\textbf{Operator} & \textbf{Task template} & \textbf{Question / supervision form} & \textbf{Purpose} \\
\midrule

\multirow{5}{0.17\linewidth}{\shortstack[l]{\textbf{Semantic}\\\textbf{Anchoring}}}
& Identification
& Describe the highlighted entity?
& Entity recognition on ERP inputs. \\

& Attribute QA
& What visual attributes does the entity have?
& Fine-grained visual semantics. \\

& Existence
& Is there a target entity matching this description?
& Entity-language grounding. \\

& Counting
& How many entities of a given type are visible?
& Global entity awareness. \\

& Scene captioning
& Describe the region concisely or densely.
& scene-level semantic description. \\

\midrule

\multirow{4}{0.17\linewidth}{\shortstack[l]{\textbf{Angular}\\\textbf{Grounding}}}
& Absolute direction
& Which direction sector contains the target entity?
& Coarse spherical localization. \\

& Angular center prediction
& Predict the target center direction $(\theta,\phi)$.
& Fine-grained spherical localization. \\

& Angular footprint prediction
& Predict the target angular region $(\theta,\phi,\Delta\theta,\Delta\phi)$.
& BFOV-style spatial grounding. \\

& Referring grounding
& Localize the entity described by the query.
& Language-to-region alignment. \\

\midrule

\multirow{4}{0.17\linewidth}{\shortstack[l]{\textbf{Spherical}\\\textbf{Reference-frame}\\\textbf{Transformation}}}
& Relative direction
& Where is entity B relative to entity A on the sphere?
& Pairwise spherical relation. \\

& Camera rotation transform
& After turning left/right by a given angle, where is the target?
& Observer heading update. \\

& Object-conditioned reorientation
& If facing entity A, where is entity B in the new frame of reference?
& Object-centered reference frame. \\

% & Seam-aware relation
% & Are two boundary regions adjacent in the real 360$^\circ$ scene?
% & Wrap-around topology awareness. \\

\midrule

\multirow{4}{0.17\linewidth}{\shortstack[l]{\textbf{Depth-aware}\\\textbf{3D Relation}}}
& Observer distance
& Which entity is closer to the observer?
& Depth comparison. \\

& Distance ordering
& Rank entities by distance from the observer.
& Global depth structure. \\

& 3D relative position
& What is the relation between target A and target B? 
& Viewer-centered 3D relation. \\

& Compound 3D relation
& Which option satisfies a combined relation such as front-left-above?
& Multi-axis spatial reasoning. \\

\bottomrule
\end{tabularx}
\end{table}

\subsection{Prompt Template}
\label{prompt}

This section details the prompt templates used in our evaluation pipeline.
The core prompt explicitly defines the input as a full-surround ERP panorama, specifies the observer-centered reference frame, and standardizes the meanings of BFOV localization, relative direction, camera rotation, object-conditioned reorientation, physical distance, and relative 3D position.
This unified formulation reduces ambiguity in spatial supervision and ensures that all task templates are grounded in the same panoramic coordinate system.

\textbf{Unified pano-native system prompt.}
We use the following system prompt as the default instruction shared across ERP-native tasks.

\begin{tcolorbox}[
    colback=gray!4,
    colframe=blue!55!black,
    title=\textbf{Unified Pano-Native System Prompt},
    fonttitle=\bfseries,
    arc=2mm,
    boxrule=0.8pt,
    left=1.5mm,
    right=1.5mm,
    top=1mm,
    bottom=1mm
]
\footnotesize
You are a multimodal assistant specialized in ERP (equirectangular projection) panoramic image understanding.

The input image is an ERP panorama representing the full 360-degree surrounding scene captured from a single fixed viewpoint. 
It should be interpreted as a continuous panoramic observation centered at the current observer, rather than as a standard perspective image.

All directional and spatial judgments are defined in an observer-centered reference frame anchored at the current observer. 
The image center corresponds to the current front direction.

BFOV is represented as $[\mathrm{yaw}, \mathrm{pitch}, x_{\mathrm{fov}}, y_{\mathrm{fov}}]$ in degrees. 
In this representation, yaw and pitch denote the center direction of the target object, while $x_{\mathrm{fov}}$ and $y_{\mathrm{fov}}$ denote the angular width and angular height covering the target object.

Positive yaw corresponds to the observer's right, and negative yaw corresponds to the observer's left. 
Positive pitch corresponds to the upward direction, and negative pitch corresponds to the downward direction. 
The valid range of yaw is $[-180^\circ, 180^\circ)$, the valid range of pitch is $[-90^\circ, 90^\circ)$, and the valid ranges of $x_{\mathrm{fov}}$ and $y_{\mathrm{fov}}$ are $(0^\circ, 180^\circ]$.

Relative direction is defined by comparing the center direction of the target object with that of the reference object while keeping the current observer orientation fixed.

Camera rotation is defined as an in-place change of observer orientation without any change in observer position. 
Under this operation, the current front direction is updated according to the specified turn angle and turn direction, and the target object is then judged in the rotated observer frame.

Object-conditioned reorientation is defined as an in-place reorientation in which the center direction of the specified facing object becomes the new front direction. 
The target object is then judged in the reoriented observer frame.

Physical distance is defined in scene 3D space relative to the current observer, namely the current camera position.

Relative 3D position is defined as the positional relation of one object to another in the current observer-centered 3D frame.

Return only the requested answer in the required format unless explicitly instructed otherwise.
\end{tcolorbox}

\textbf{Additional ERP guidance for prompt-based inference.}
For the prompt-only ERP baselines, we further provide explicit guidance to help the model interpret ERP panoramas.
We consider two forms of pano-specific guidance: a textual reference appendix, which verbally explains the ERP coordinate layout, and a visual guidance appendix, which combines an overlaid coordinate grid with accompanying text instructions.
These auxiliary prompts are designed to teach the model how to read ERP panoramas, rather than to provide task-specific reasoning rules.

\textbf{Text-only ERP reference.}
The following appendix provides a concise natural-language explanation of the ERP reference system.

\begin{tcolorbox}[
    colback=gray!4,
    colframe=orange!70!black,
    title=\textbf{ERP Reference System Appendix},
    fonttitle=\bfseries,
    arc=2mm,
    boxrule=0.8pt,
    left=1.5mm,
    right=1.5mm,
    top=1mm,
    bottom=1mm
]
\footnotesize
Reference System (Equirectangular Projection):
\begin{itemize}\setlength{\itemsep}{1pt}
    \item Image center: yaw $0^\circ$, corresponding to the front direction.
    \item Left and right image boundaries: yaw $\pm 180^\circ$, corresponding to the back direction.
    \item Yaw $90^\circ$: one quarter of the image width to the right of the center.
    \item Yaw $-90^\circ$: one quarter of the image width to the left of the center.
    \item Vertical axis: pitch $0^\circ$ is the horizon, pitch $90^\circ$ is the top (zenith), and pitch $-90^\circ$ is the bottom (nadir).
\end{itemize}
\end{tcolorbox}

\textbf{Visual ERP guidance.}
We also construct a visual prompt by overlaying a coordinate grid on the ERP panorama and appending a short textual explanation of the grid semantics.
This visual guidance makes the yaw--pitch structure directly observable in the image.
\textbf{Grid rendering for visual prompting.}
To generate the visual prompt, we render yaw grid lines every $30^\circ$ and pitch grid lines every $15^\circ$, and mark the image center with a yellow crosshair corresponding to $(0^\circ,0^\circ)$.
The yaw lines are drawn in green and the pitch lines in blue, each with numerical labels placed along the image borders.
This produces a visually interpretable ERP coordinate frame that can be directly consumed by the model.

\begin{tcolorbox}[
    colback=gray!4,
    colframe=green!50!black,
    title=\textbf{ERP Visual Guidance Appendix},
    fonttitle=\bfseries,
    arc=2mm,
    boxrule=0.8pt,
    left=1.5mm,
    right=1.5mm,
    top=1mm,
    bottom=1mm
]
\footnotesize
Visual Guidance System:
\begin{itemize}\setlength{\itemsep}{1pt}
    \item The image is overlaid with a coordinate grid and numerical labels.
    \item Green vertical lines represent yaw angles. Numerical labels (e.g., $-180,-150,\ldots,0,\ldots,180$) are shown at the top and bottom.
    \item Blue horizontal lines represent pitch angles. Numerical labels (e.g., $-90,-75,\ldots,0,\ldots,90$) are shown at the left and right ends.
    \item A yellow crosshair marks the front direction at $(0^\circ,0^\circ)$.
\end{itemize}

Task: Use the visual grid lines and numerical labels as a ruler to estimate the target center direction $[\mathrm{yaw},\mathrm{pitch}]$. Interpolate between lines if the target does not lie exactly on a grid intersection.
\end{tcolorbox}

\section{Benchmark Setting}
\label{app:benchmark_setting}

\subsection{PanoSpace-Bench}

% \section{PanoSpace-Bench}
% \label{sec:benchmark}

% \input{table/table_benchmark}

We introduce \textbf{PanoSpace-Bench}, a diagnostic benchmark for evaluating whether MLLMs understand equirectangular panoramas as a continuous, observer-centered representation of omnidirectional 3D spaces.
Unlike existing panoramic benchmarks that typically focus on individual tasks such as VQA, captioning, grounding, or navigation, PanoSpace-Bench is designed to probe the capability structure behind first-person 360$^\circ$ spatial understanding, where directions, reference frames, and 3D relations are all defined around the observer.

\textbf{Data separation.}
To avoid data leakage, PanoSpace-Bench is constructed from image sources that are completely separate from those used for the ERP-native instruction-tuning corpus.
Specifically, benchmark panoramas are collected from different Internet sources, deduplicated against the training corpus, and manually verified for image quality and valid ERP layout.
In addition to image-level separation, the benchmark questions are also separated from the training questions.
Although some benchmark categories correspond to the same high-level abilities as the training tasks, we design distinct question formats and evaluation protocols for PanoSpace-Bench.
Therefore, the benchmark evaluates generalization to unseen panorama sources and task formulations, rather than memorization of training images or generated QA templates.

As shown in Table~\ref{tab:benchmark}, PanoSpace-Bench contains four ability families. 
Panoramic localization evaluates whether a model can ground targets in yaw--pitch space, including coarse absolute-direction classification and fine-grained BFOV localization. 
Spherical relational reasoning evaluates object-object angular relations and egocentric reference-frame transformations, including relative direction, camera rotation, and object-conditioned reorientation. 
Omnidirectional 3D spatial reasoning evaluates distance comparison and relative 3D position reasoning in the surrounding scene. 
ERP representation properties focuses on projection-specific topology, instantiated by seam-continuity questions that test whether the model treats the left and right ERP borders as adjacent on the viewing sphere.

Two design choices distinguish PanoSpace-Bench from prior omnidirectional evaluations. 
First, the benchmark emphasizes ERP-specific spatial modes rather than generic QA over panoramic content. 
Second, it separates semantic recognition from spatial correctness: a model may correctly identify the objects in a question but still fail if it does not understand their positions and relations in the observer-centered spherical frame. 
We report category-wise multiple-choice accuracy for closed-form tasks and BFOV mean IoU for fine-grained angular localization.

PanoSpace-Bench contains four ability families and eight task categories, with 250 questions for each category and 2,000 questions in total.
Except for BFOV localization, all tasks are formulated as multiple-choice questions and evaluated by exact choice accuracy.
BFOV localization requires the model to predict an angular bounding field-of-view in the format $[\mathrm{yaw}, \mathrm{pitch}, x_{\mathrm{fov}}, y_{\mathrm{fov}}]$, and is evaluated by angular IoU between the predicted and ground-truth BFOV regions.

\textbf{Multiple-choice evaluation.}
For each multiple-choice question, the model is required to output one option from the candidate set.
We parse the predicted response into a choice label $\hat{y}_i$ and compare it with the ground-truth label $y_i$.
Invalid or unparsable responses are counted as incorrect.
For a task category with $N$ examples, the accuracy is computed as
\begin{equation}
    \mathrm{Acc}
    =
    \frac{1}{N}
    \sum_{i=1}^{N}
    \mathbb{I}(\hat{y}_i = y_i),
\end{equation}
where $\mathbb{I}(\cdot)$ is the indicator function.

\textbf{BFOV localization evaluation.}
For BFOV localization, the model predicts an angular region
\begin{equation}
    \hat{b}_i =
    [\hat{\theta}_i, \hat{\phi}_i, \hat{w}_i, \hat{h}_i],
\end{equation}
where $(\hat{\theta}_i,\hat{\phi}_i)$ is the predicted yaw-pitch center and $(\hat{w}_i,\hat{h}_i)$ are the predicted horizontal and vertical angular extents.
The ground-truth BFOV is denoted as
\begin{equation}
    b_i =
    [\theta_i, \phi_i, w_i, h_i].
\end{equation}
Each BFOV is converted to an angular rectangle on the ERP sphere:
\begin{equation}
    R(b_i)
    =
    \left[
    \theta_i-\frac{w_i}{2},
    \theta_i+\frac{w_i}{2}
    \right]
    \times
    \left[
    \phi_i-\frac{h_i}{2},
    \phi_i+\frac{h_i}{2}
    \right].
\end{equation}
We then compute angular IoU as
\begin{equation}
    \mathrm{IoU}(\hat{b}_i,b_i)
    =
    \frac{
    |R(\hat{b}_i) \cap R(b_i)|
    }{
    |R(\hat{b}_i) \cup R(b_i)|
    },
\end{equation}
where $|\cdot|$ denotes the angular area on the yaw-pitch domain.
The final BFOV localization score is the mean IoU over all BFOV examples:
\begin{equation}
    \mathrm{mIoU}
    =
    \frac{1}{N}
    \sum_{i=1}^{N}
    \mathrm{IoU}(\hat{b}_i,b_i).
\end{equation}

When computing yaw overlap, we account for the circular wrap-around of ERP panoramas at the left-right boundary.
Predictions with invalid formats or out-of-range angular values are treated as invalid and assigned zero IoU.

\begin{table*}[t]
\centering
\small
\setlength{\tabcolsep}{4.2pt}
\renewcommand{\arraystretch}{1.15}
\caption{
Taxonomy of PanoSpace-Bench. 
The benchmark is organized into four pano-centered ability families and eight diagnostic task categories.
}
\label{tab:benchmark}
\begin{tabular}{p{3.0cm} p{2.9cm} p{5.8cm} p{1.2cm}}
\toprule
Family & Category & Representative question form & Metric \\
\midrule

Panoramic localization
& Absolute direction
& Where is the target object located relative to the observer: left-back, left, or right ...?
& MC acc. \\

Panoramic localization
& BFOV localization
& What BFOV $[\mathrm{yaw}, \mathrm{pitch}, x_{\mathrm{fov}}, y_{\mathrm{fov}}]$ localizes the target object in the panorama?
& mIoU \\

\midrule

Spherical relation
& Relative direction
& Where is object A relative to object B on the viewing sphere?
& MC acc. \\

Spherical relation
& Camera rotation
& After the observer turns by a specified yaw angle, where would the target object appear?
& MC acc. \\

Spherical relation
& Object reorientation
& If the observer first faces the reference object, where is the target object?
& MC acc. \\

\midrule

3D spatial reasoning
& Observer distance
& Which listed object is physically closest to the observer in 3D space?
& MC acc. \\

3D spatial reasoning
& Relative 3D position
& Which option best describes the target object's 3D position relative to the observer, combining direction and depth?
& MC acc. \\

\midrule

ERP property
& Seam continuity
& For target A near the right boundary of the ERP panorama, which listed object is nearest to it in the full 360 scene?
& MC acc. \\

\bottomrule
\end{tabular}
\end{table*}

% \begin{table*}[t]
% \centering
% \small
% \setlength{\tabcolsep}{4.2pt}
% \renewcommand{\arraystretch}{1.15}
% \caption{
% Taxonomy of PanoSpace-Bench. 
% The benchmark is organized into four pano-centered ability families and eight diagnostic task categories.
% }
% \label{tab:benchmark}
% \begin{tabular}{p{3.0cm} p{2.9cm} p{5.8cm} p{1.2cm}}
% \toprule
% Family & Category & Diagnostic focus & Metric \\
% \midrule

% Panoramic localization
% & Absolute direction
% & Coarse localization of a target entity in the observer-centered yaw direction.
% & MC acc. \\

% Panoramic localization
% & BFOV localization
% & Fine-grained angular grounding of a target using $[\mathrm{yaw}, \mathrm{pitch}, x_{\mathrm{fov}}, y_{\mathrm{fov}}]$.
% & mIoU \\

% \midrule

% Spherical relation
% & Relative direction
% & Object-object angular relation on the viewing sphere.
% & MC acc. \\

% Spherical relation
% & Camera rotation
% & Direction update after an egocentric heading change.
% & MC acc. \\

% Spherical relation
% & Object reorientation
% & Direction reasoning after facing a reference object.
% & MC acc. \\

% \midrule

% 3D spatial reasoning
% & Observer distance
% & Near/far relation with respect to the observer.
% & MC acc. \\

% 3D spatial reasoning
% & Relative 3D position
% & Viewer-centered relation combining direction and depth cues.
% & MC acc. \\

% \midrule

% ERP property
% & Seam continuity
% & Wrap-around topology between the left and right ERP borders.
% & MC acc. \\

% \bottomrule
% \end{tabular}
% \end{table*}

\subsection{Human-centric visual search.}
H$^\ast$Bench~\citep{yu2025thinking360} evaluates human-centric visual search in 360$^\circ$ panoramas, including Humanoid Object Search (HOS) and Humanoid Path Search (HPS). 
The original setting uses an interactive perspective-view protocol, where the model observes a local FoV, iteratively performs rotation actions, and finally submits a target direction. 
This protocol is closely related to recent ``thinking with images'' studies, which encourage vision-language models to actively ground reasoning in visual evidence~\citep{wang2025pixel, wang2025illusionintentionvisualrationale}. 
In our ERP-input setting, the model directly receives the full ERP panorama and predicts the target direction in one step without decomposing the scene into local views.
Following the official evaluation protocol of Thinking in 360, the model output is parsed as a yaw-pitch direction $(\hat{\theta},\hat{\phi})$, and a prediction is considered successful if the submitted direction falls within the annotated target region or the task-specific angular tolerance around the ground-truth direction.
We report the overall success rate as well as separate success rates for HOS and HPS:
\begin{equation}
    \mathrm{Success}
    =
    \frac{1}{N}
    \sum_{i=1}^{N}
    \mathbb{I}\!\left[
    \mathrm{hit}((\hat{\theta}_i,\hat{\phi}_i), \mathcal{T}_i)
    \right],
\end{equation}
where $\mathcal{T}_i$ denotes the ground-truth target region or valid angular tolerance, for example, $i$, and $\mathrm{hit}(\cdot)$ follows the benchmark evaluator.
For setting the tolerance interval, we follow the parameters specified in the original paper.
For perspective-view baselines, we additionally report the average number of interaction steps and model calls required before submission, following the original interactive setting.

\begin{table*}[t]
\centering
\small
\caption{
Transfer evaluation on H*Bench.
\textbf{Left:} Representative results under the original perspective-view setting and our PanoWorld zero-shot setting.
\textbf{Right:} Route-level comparison of ERP baselines, prompt-only ERP formulation, and training-based ERP adaptation.
}
\begin{subtable}[t]{0.41\textwidth}
\centering
\scriptsize
\setlength{\tabcolsep}{5.2pt}
\renewcommand{\arraystretch}{1.10}
% \caption{Main transfer comparison.}
\caption{Perspective-based methods}
\resizebox{\linewidth}{!}{
\begin{tabular}{lccc}
\toprule
Method & Overall & HOS & HPS \\
\midrule
% \multicolumn{4}{l}{Perspective-view input setting} \\
% \midrule
GPT-4o              & 21.3 & 19.7 & 23.6 \\
Gemini-2.5-Pro      & 32.3 & 31.9 & \underline{33.0} \\
Kimi-VL-A3B   & 4.6  & 4.92  & 4.32  \\
InternVL3.5-4B      & 3.8  & 3.2  & 4.8  \\
InternVL3.5-8B      & 6.7  & 6.4  & 7.2  \\
Qwen2.5-VL-3B & 11.4 & 14.8 & 6.4  \\
Qwen2.5-VL-7B & 9.3  & 11.3 & 6.3  \\
Qwen3-VL-4B & 17.5  & 19.5 & 14.4  \\
Qwen3-VL-8B & 19.1  & 23.6 & 12.2  \\
Qwen3.5-9B      & 18.9  & 21.8  & 14.5 \\
% Gemma-3-4B       & 16.05 & 17.13 & 14.44 \\
Gemma-3-12B      & 11.9 & 10.2 & 14.5 \\
HVS-3B*              & \underline{38.4} & \underline{47.3} & 24.9 \\
% \midrule
% \multicolumn{4}{l}{ERP panorama input setting} \\
\midrule
% Qwen3.5 naive H*Bench SFT   & 17.80 & 11.17 & 27.75 \\
Ours + zero-shot   & \textbf{56.1} & \textbf{61.8} & \textbf{47.5} \\
% Ours Pano-native + H*Bench FT
                            % & \textbf{70.00} & \textbf{73.00} & \textbf{64.00} \\
\bottomrule
\end{tabular}
}
\label{tab:thinking360_main_sup}
\end{subtable}
\hfill
\begin{subtable}[t]{0.54\textwidth}
\centering
\scriptsize
\setlength{\tabcolsep}{2.7pt}
\renewcommand{\arraystretch}{1.12}
% \caption{Route-level comparison.}
\caption{ERP panorama-based methods}
\resizebox{\linewidth}{!}{
\begin{tabular}{lccccc}
\toprule
Route / Setting & Overall & HOS & HPS & Yaw Acc & Pitch Acc \\
\midrule
% \multicolumn{6}{l}{Direct ERP inference baselines} \\
% \midrule
GPT-4o             & 30.1   & 39.1   & 17.1   & 38.5   & 64.2 \\
Gemini-2.5-Pro     & \underline{46.9}   & \underline{55.3}   & \underline{34.3}   & \underline{52.5}   & \underline{71.6} \\
InternVL3.5-4B     & 11.6   & 12.8   & 9.75   & 22.1   & 41.5 \\
InternVL3.5-8B     & 14.9   & 18.0   & 10.25   & 19.2   & 38.5 \\
Qwen2.5-VL-7B      & 9.1   & 9.6   & 8.5   & 11.9   & 32.3 \\
Qwen3-VL-4B      & 12.8   & 14.3   & 10.5   & 15.6   & 40.0 \\
Qwen3-VL-8B        & 13.1   & 15.0   & 10.3   & 16.8   & 39.2 \\
% Qwen3.5-27B         & 21.9 & 27.5 & 9.8 & 26.2  & 48.9 \\
Gemma-3-12B         & 9.6   & 10.9   & 7.5   & 14.0   & 40.8 \\
\midrule
% \multicolumn{6}{l}{Prompt-only ERP route} \\
% \midrule
% Direct output           & 19.4 & 26.2 & 9.3 & 23.5   & 46.5 \\
Qwen3.5-9B         & 19.4 & 26.2 & 9.3 & 23.5   & 46.5 \\
+ Text prompt             & 38.5 & 43.3 & 31.2 & 40.0 & 49.5 \\
+ Visual prompt      & 40.4 & 46.0 & 32.0 & 43.5 & 52.0 \\
% Detect + rule$^\dagger$ & \textbf{72.10} & \textbf{88.33} & 47.75 & xx & xx \\
\midrule
% \multicolumn{6}{l}{Training-based ERP route} \\
% \midrule
Qwen3.5 + H* SFT       & 17.8 & 11.1 & 27.7 & 25.3 & 42.5 \\
% ERP-native zero-shot    & 56.10 & 61.83 & 47.50 & 64.60 & 77.00 \\
Ours + H* SFT & \textbf{70.1} & \textbf{73.1} & \textbf{64.2} & \textbf{74.1} & \textbf{85.5} \\
\bottomrule
\end{tabular}
}
\label{tab:thinking360_routes_sup}
\end{subtable}
\vspace{-0.5em}
\label{tab:thinking360_transfer_full}
\vspace{-0.5em}
\end{table*}

\subsection{R2R-CE vision-and-language navigation.}
We further evaluate transfer to embodied navigation on R2R-CE Val-Unseen. 
Unlike conventional VLN methods that often use panoramas to construct candidate perspective views or rely on additional observations such as odometry, depth, or single-view RGB streams, our model directly takes the ERP panorama as the visual observation and predicts the navigation direction from the full-surround input.
We follow the standard R2R-CE evaluation protocol and report navigation error (NE), oracle success rate (OSR), success rate (SR) and success weighted by path length (SPL).
For an episode $i$, NE is the final geodesic distance $d_i$ between the agent and the goal. 
SR is computed with the standard success threshold $\tau=3$m:
\begin{equation}
    \mathrm{SR}
    =
    \frac{1}{N}
    \sum_{i=1}^{N}
    \mathbb{I}(d_i \leq \tau).
\end{equation}
OSR uses the closest distance to the goal along the trajectory instead of the final distance:
\begin{equation}
    \mathrm{OSR}
    =
    \frac{1}{N}
    \sum_{i=1}^{N}
    \mathbb{I}\!\left(\min_{t} d_{i,t} \leq \tau\right).
\end{equation}
SPL additionally penalizes unnecessarily long paths:
\begin{equation}
    \mathrm{SPL}
    =
    \frac{1}{N}
    \sum_{i=1}^{N}
    S_i
    \frac{\ell_i}{\max(p_i,\ell_i)},
\end{equation}
where $S_i$ is the success indicator, $\ell_i$ is the shortest-path length, and $p_i$ is the executed path length.
For fair comparison with recent RGB/video-based methods, our VLN fine-tuning uses only the R2R and RxR training sets, and PanoWorld uses only 80\% of the training data.

\FloatBarrier
\begin{figure*}[!t]
    \centering
    \captionsetup{font=footnotesize,skip=4pt}

    \begin{minipage}[t]{0.18\textwidth}
        \centering
        \includegraphics[width=\linewidth]{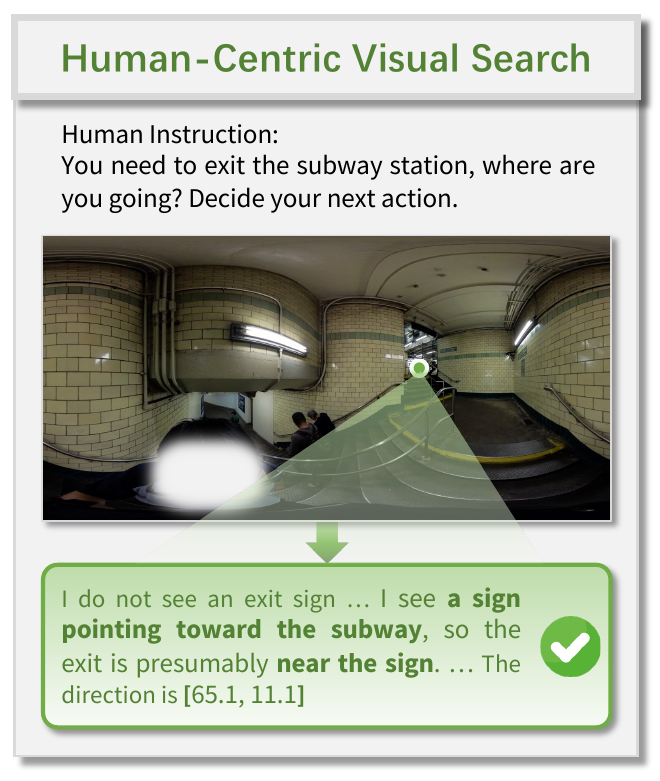}\\[-0.15em]
        {\scriptsize (a) Visual search}
    \end{minipage}
    \hfill
    \begin{minipage}[t]{0.18\textwidth}
        \centering
        \includegraphics[width=\linewidth]{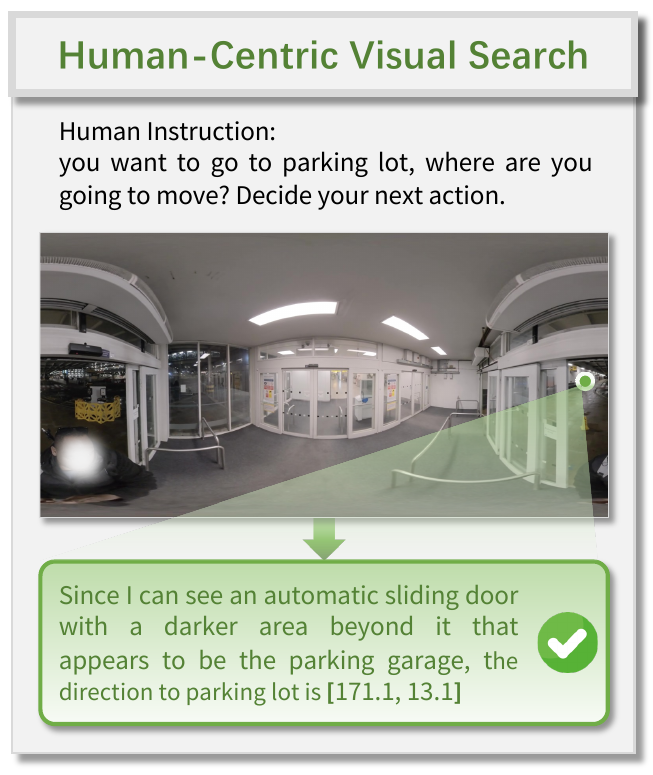}\\[-0.15em]
        {\scriptsize (b) Visual search}
    \end{minipage}
    \hfill
    \begin{minipage}[t]{0.18\textwidth}
        \centering
        \includegraphics[width=\linewidth]{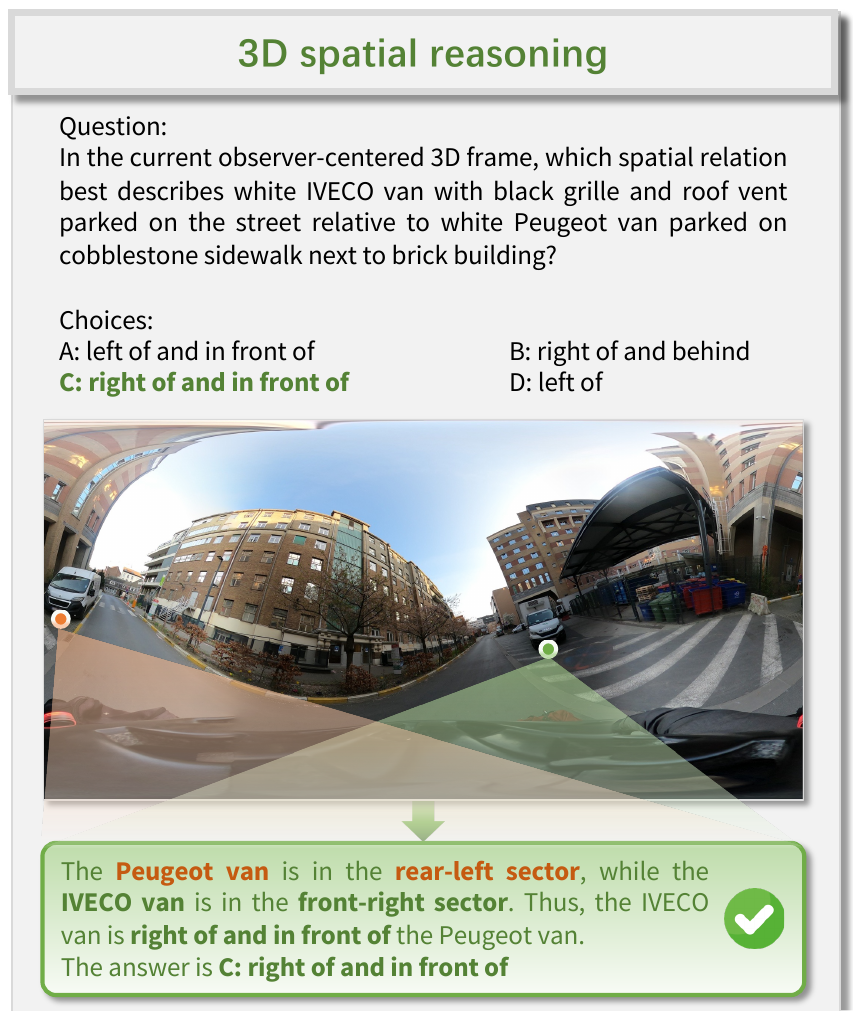}\\[-0.15em]
        {\scriptsize (c) 3D relation}
    \end{minipage}
    \hfill
    \begin{minipage}[t]{0.16\textwidth}
        \centering
        \includegraphics[width=\linewidth]{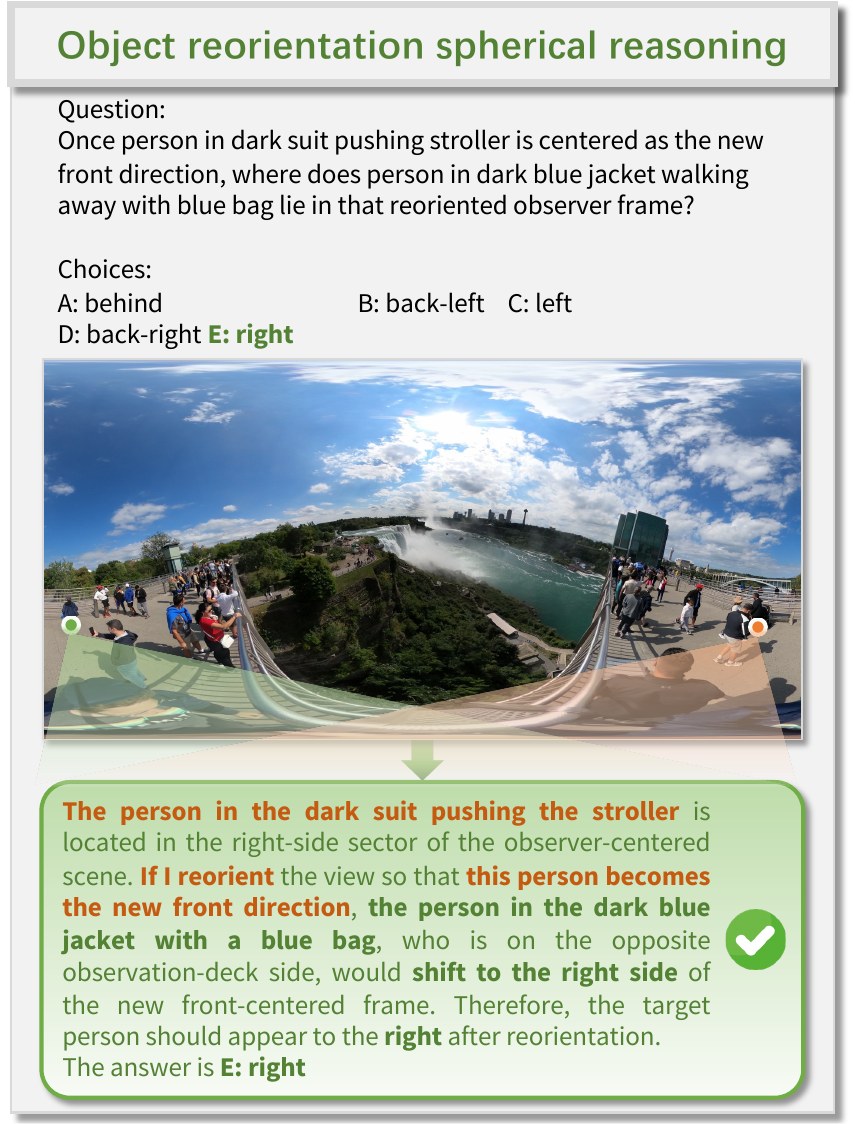}\\[-0.15em]
        {\scriptsize (d) Reorientation}
    \end{minipage}
    \hfill
    \begin{minipage}[t]{0.18\textwidth}
        \centering
        \includegraphics[width=\linewidth]{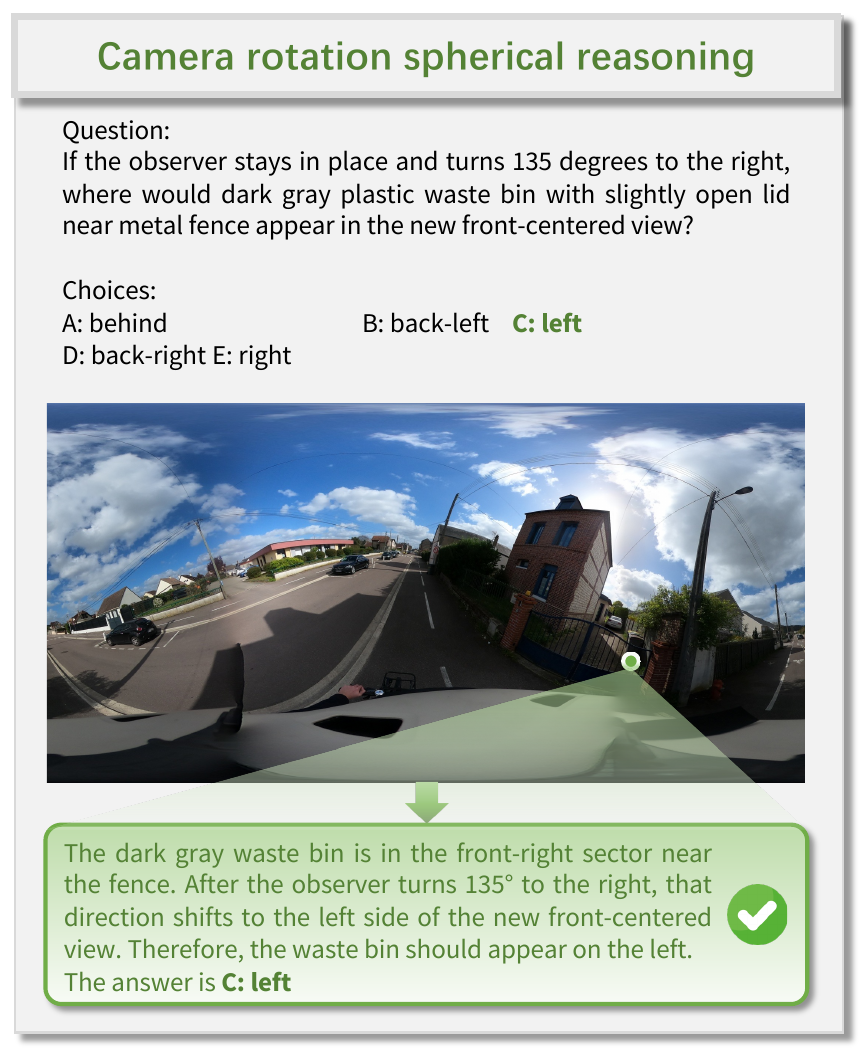}\\[-0.15em]
        {\scriptsize (e) Camera rotation}
    \end{minipage}

    \vspace{0.35em}
    \caption{
    Case studies of pano-native spatial reasoning.
    The first two examples show downstream human-centric visual search, while the remaining examples show representative PanoSpace-Bench tasks covering 3D relation, object-conditioned reorientation, and camera-rotation reasoning.
    }
    \label{fig:case_study_gallery}
    \vspace{-0.3em}
\end{figure*}
\FloatBarrier

% \FloatBarrier
\section{Case Study}
\label{Case_study}

We present qualitative examples for both downstream transfer and diagnostic evaluation.
Figure~\ref{fig:case_study_gallery} (a--b) shows human-centric visual search cases, where the model directly reasons over a full ERP panorama to infer the target movement direction.
These examples illustrate that pano-native spatial learning supports practical 360$^\circ$ search without decomposing the scene into local perspective views. Moreover, as shon in Table~\ref{tab:thinking360_transfer_full}, we present more comprehensive data results on H*.

Figure~\ref{fig:case_study_gallery} (c--e) shows representative PanoSpace-Bench cases, including 3D spatial relations, object-conditioned reorientation, and camera-rotation reasoning.
These examples further demonstrate that the learned representation supports controlled evaluation of observer-centered spherical reasoning.

\vspace{-0.5em}
\section{Efficiency Study}
\vspace{-0.5em}
\label{efficiency_study}

We compare the inference efficiency of direct ERP reasoning with the perspective-view rotation paradigm on H$^\ast$ tasks.
Rotation-based methods observe only one local FoV at each step and therefore require multiple sequential model calls before submitting the final direction.
In contrast, our model directly consumes the full ERP panorama and predicts the answer in a single forward pass.

\begin{table}[t]
\centering
\small
\caption{
Efficiency comparison between perspective-view rotation and direct ERP inference on H$^\ast$ tasks. Direct ERP inference is more efficient in terms of interaction steps, sequential decision cost, and global spatial coverage.
}
\setlength{\tabcolsep}{4pt}
\begin{tabular}{lcccccc}
\toprule
Method & Input & Res. & Steps $\downarrow$ & Calls $\downarrow$ & Eff. Input Tokens & Rel. Cost \\
\midrule
Qwen3-VL-4B & Persp. rotation & $720^2$ & 6.27 & 6.27 & 29.6k & 1.80$\times$ \\
Qwen3-VL-8B & Persp. rotation & $720^2$ & 6.34 & 6.34 & 29.9k & 1.81$\times$ \\
H$^\ast$-trained Qwen2.5-VL & Persp. rotation & $720^2$ & 3.70 & 3.70 & 19.2k & 1.16$\times$ \\
H$^\ast$-trained Qwen3-VL & Persp. rotation & $720^2$ & 3.58 & 3.58 & 18.7k & 1.13$\times$ \\
\midrule
Ours & Direct ERP & $1600{\times}800$ & 1.00 & 1.00 & 16.5k & 1.00$\times$ \\
\bottomrule
\end{tabular}
\label{tab:erp_vs_rotation_efficiency}
\end{table}

As shown in Table~\ref{tab:erp_vs_rotation_efficiency}, perspective-view rotation requires 3.58--6.34 interaction steps on average, leading to 18.7K--29.9K effective input tokens.
This corresponds to 1.13--1.81$\times$ the cost of direct ERP inference.
Our method requires only one step and one model call, with 16.5K effective input tokens, while maintaining full 360$^\circ$ spatial coverage.
This demonstrates that pano-native spatial learning replaces iterative local-view search with a unified and efficient full-surround inference paradigm.

% \vspace{-0.8em}
\section{Ablation study}
% \vspace{-0.8em}
\label{sec:appendix_ablation}

% \input{table/pipeline_ablation}

% \textbf{Metadata verification.}
% Table~\ref{tab:pipeline_ablation} shows that both verification modules are important for reliable ERP supervision. Starting from the unverified baseline, detection verification improves overall accuracy from 38.8 to 46.4 by filtering geometrically unstable proposals, while semantic verification raises it to 48.0 by removing inconsistent language-region pairs. Combining both yields the best result of 55.1, with gains across localization, directional reasoning, 3D reasoning, and seam continuity. This confirms that data quality is a major factor in pano-native learning.

% \input{table/trainable_ablation}

% \textbf{Trainable component ablation.} Table~\ref{tab:trainable_ablation} examines the trainable scope during pano-native adaptation. We compare different update strategies over the vision encoder, the vision-language interface, and the language model to identify where pano-native spatial learning mainly takes place. The results indicate that panoramic reasoning depends not only on language-side adaptation, but also on updating the visual and cross-modal components that encode and align ERP geometry.

% \input{table/arch_ablation}

\textbf{Architecture ablation.}
We ablate two key design choices of the pano-aware adapter: the fusion mechanism and the insertion position.
For the fusion mechanism, we compare simple residual fusion, which directly adds projected spherical features to visual tokens, with cross-attention fusion, where visual tokens adaptively attend to spherical spatial tokens.
For the insertion position, we inject the spherical adapter at three stages of the visual stream: immediately after patch embedding, after visual token merging, and after the visual encoder output.
This allows us to evaluate whether ERP geometry should be introduced early at the patch level or later after visual abstraction.

% Table~\ref{tab:arch_ablation} compares residual fusion and cross-attention at different insertion positions. Patch-level cross-attention performs best overall, improving accuracy from 0.484 to 0.551 and yielding the strongest spherical relation average (0.460) and 3D spatial average (0.488). Residual fusion also helps in some settings, especially seam continuity, but is less consistent on relation-heavy categories. These results support the proposed SSCA design: geometry is most effective when injected early and through content-dependent interaction with visual tokens.

\vspace{-0.5em}
\section{Limitations}
\vspace{-0.5em}
\label{app:limitations}

While PanoWorld demonstrates strong pano-native spatial reasoning, several limitations remain.
First, our metadata construction pipeline relies on automatic open-world detection, MLLM-based semantic annotation, referring re-detection, and panoramic depth estimation.
Although we introduce two-level verification to improve reliability, errors from these components may still propagate to the final metadata graph.
Second, PanoSpace-Bench is designed as a diagnostic benchmark for observer-centered ERP spatial reasoning, and therefore does not cover all possible panoramic tasks, such as long-horizon embodied interaction, dynamic scenes, or multi-agent navigation.

These limitations suggest several directions for future work.
On the data side, more reliable panoramic perception modules and stronger cross-modal verification could further improve metadata quality.
On the evaluation side, extending pano-native benchmarks from static ERP reasoning to interactive navigation, temporal panoramic videos, and dynamic 3D environments would provide a broader testbed for full-surround spatial intelligence.
We hope PanoWorld and PanoSpace-Bench provide a foundation for these future studies.

%%%%%%%%%%%%%%%%%%%%%%%%%%%%%%%%%%%%%%%%%%%%%%%%%%%%%%%%%%%%

% \newpage
% \input{checklist.tex}

\end{document}